\documentclass[preprint,12pt,authoryear]{elsarticle}

\usepackage{amssymb}
\usepackage{todonotes}
\usepackage{caption}
\usepackage{subcaption}
\usepackage{xcolor}
\definecolor{greenf}{HTML}{48bf91}
\definecolor{bluef}{HTML}{00a1df}
\usepackage{graphicx}
\usepackage{amsmath}
\usepackage{amsfonts}
\usepackage{algorithmic}
\usepackage{makecell}
\usepackage{multirow}
\DeclareUnicodeCharacter{2212}{-} \usepackage[ruled,vlined,linesnumbered]{algorithm2e}
\usepackage{amsthm}
\usepackage{hyperref}

\newtheorem{definition}{Definition}[section]  
\newtheorem{problem}{Problem}
\newcommand{\A}{\mathcal{A}}
\newcommand{\model}{\texttt{NICKI}}

\journal{Neural Networks}

\begin{document}

\begin{frontmatter}



\title{Node Injection for Class-specific Network Poisoning}


\author[label1]{Ansh Kumar Sharma\fnref{label3}}
\author[label1]{Rahul Kukreja\fnref{label3}}
\author[label1]{Mayank Kharbanda\fnref{label3}}
\author[label2]{Tanmoy Chakraborty}
\affiliation[label1]{organization={Indraprastha Institute of Information Technology},
            state={Delhi},
            country={India}}
\affiliation[label2]{organization={Indian Institute of Technology},
            state={Delhi},
            country={India}}
     \fntext[label3]{First three authors contributed equally.}       

\begin{abstract}
Graph Neural Networks (GNNs) are powerful in learning rich network representations that aid the performance of  downstream tasks. However, recent studies showed that GNNs are vulnerable to adversarial attacks involving node injection and network perturbation. Among these, node injection attacks are more practical as they don't require manipulation in the existing network and can be performed more realistically. 
In this paper, we propose a novel problem statement -- a {\em class-specific poison attack} on graphs in which the attacker aims to misclassify specific  nodes in the target class into a different class using node injection. Additionally,  nodes are injected in such a way that they camouflage as benign nodes. 
We propose \model, a novel attacking strategy that utilises an optimization-based approach to sabotage the performance of GNN-based node classifiers. \model\ works in two phases -- it first learns the node representation and then generates the features and edges of the injected nodes. Extensive experiments and ablation studies on four benchmark networks show that \model\ is consistently better than four baseline attacking strategies for misclassifying nodes in the target class. We also show that the injected nodes are properly camouflaged as benign, thus making the poisoned graph indistinguishable from its clean version w.r.t various topological properties.
\end{abstract}




\end{frontmatter}




\section{Introduction}
The network is often used as a simple abstraction of a complex system to solve various real-world tasks such as node classification \citep{kipf2016semi, hamilton2017inductive}, link prediction \citep{kipf2016variational} and community detection \citep{jin2019graph}. Recent studies have mainly focused on obtaining a rich  representation of networks using graph representation learning (GRL) that further improves the performance of the downstream tasks. Of late, graph neural networks (GNNs) have dominated the literature of GRL by showing remarkable performance in tasks such as node classification. With the increasing usage of GNNs, another body of literature \citep{zugner2018adversarial, dai2018adversarial, sun2020adversarial} alerted their limitations by showing how they are vulnerable to adversarial attacks on graphs.

{\bf Adversarial attacks on graphs.} Adversarial attacks on graph embedding algorithms are challenging due to their discrete and combinatorial nature \citep{bojchevski2019adversarial}. The majority of studies on adversarial graph attacks assume that the attacker gains access to an existing node and then manipulates its adjacent edges \citep{zugner2018adversarial, dai2018adversarial}. If the network is attributed, then the features of the corresponding nodes are also manipulated. The attack is generally carried out in a restricted setting, limiting the number of perturbations allowed. A few studies argued that such an attacking strategy is infeasible as it requires a high authority to compromise an existing node (e.g., a user account in a social network)  \citep{zugner2018adversarial}. Recent studies introduced a new setting wherein the attacker injects new nodes into the network and performs network perturbations restricted to them \citep{wang2018attack, sun2020adversarial}, which is more realistic as the attacker would have complete command on the injected nodes. For instance, injecting a new node into social networks is equivalent to creating a fake user profile, which is comparatively easier than compromising an existing user account. 

{\bf Evasion vs poison attacks.} Graph adversarial attacks can be classified into two broad categories -- evasion attack and poison attack. In an evasion attack \citep{dai2018adversarial}, the attack takes place on the test data; this is in contrast to the poison attack, which makes changes to the training data \citep{zugner2019adversarial, bojchevski2019adversarial}. Once poisoned, the graph is trained again, which requires solving a bi-level optimization function, making the poison attack more difficult than the evasion attack. Due to the dynamic nature of networks like social networks and transaction networks, graph poisoning attack seems realistic and challenging since these networks require models to be retrained more frequently as new data appears over time.

{\bf Contribution I: \textit{Class-specific network poisoning} --  A novel problem statement.}
Our attack aims to misclassify specific \textit{target class} nodes into a different class
using node injection. We call this different class our \textit{base class}. 
Many real-world scenarios back such an attack -- (i) Misguiding any fraud detection algorithm in a two-class network of transactions with classes denoting `fraudulent' and `legitimate' users. The goal of the attacker would be to get fraudulent users labelled as legitimate. (ii) Misclassifying low credit users to high credit ones in a network where user relationships are leveraged for credit risk assessment. (iii) {\color{black}Injection of spies by a counter-terrorism squad into a network of `terrorists', `civilians' and `soldiers' so as to baffle the node classifier in identifying the spies in the `terrorist' target class.}

\citet{shafahi2018poison} defined a similar problem statement for the image domain which works only for continuous spaces. However, we overcome the challenge of discrete space in graph and introduce a novel optimisation based method.
During  network poisoning, it is equally important that an injected node and an existing node look alike so that the defender won't trivially spot the attacker node. To address this, we also take measures to ensure that the injected nodes replicate both the topological structure as well as the features of the nodes present in the {\em base class}. This eventually helps in polarising the target nodes into the \textit{base class}. Continuing with the above examples, in the transaction network, the attacking users will be disguised as legitimate users (base class) and make the model misclassify fraudulent users as legitimate. 

{\bf Contribution II: \model\ -- A novel graph poisoning model.}
We propose \model, an optimization-based node injection approach to hinder the \textit{class-specific network poisonoing} task within a budget constraint.
We further provide a method to hide the attacker nodes under the hood of base class so as to confuse the annotator and the node classifier

{\bf Contribution III: Extensive evaluation.}
We demonstrate the efficacy of \model\ on  different datasets -- discrete and continuously attributed, small and medium scale. 
\model\ outperforms four network poisoning methods with significant margins with up to $42\%$ drop in node classification accuracy even with tighter constraints.
We empirically show the working of our hiding module by studying the generated features and edges, and evaluating them under robust models and homophily defenders.

{\color{black}{\bf Reproducibility.} The source codes and datasets and other execution-related information can be found at \href{https://github.com/rahulk207/nicki}{https://github.com/rahulk207/nicki}.} 

\section{Related Work}
In general, deep learning algorithms have been shown to be vulnerable to adversarial attacks \citep{goodfellow2014explaining, jia2017adversarial}. In graph learning,  GNNs have achieved fascinating results in a variety of graph-based tasks, such as node classification \citep{kipf2016semi, hamilton2017inductive, velivckovic2017graph, xu2019graph, xu2020graph}, drug design \citep{jiang2020drug} and social recommendation \citep{ying2018graph, fan2019graph}. Of late, various studies \citep{zugner2018adversarial, dai2018adversarial, zugner2019adversarial, bojchevski2019adversarial, sun2020adversarial, wang2020scalable} pointed out that GNNs too are susceptible to adversarial attacks, such as their counterparts in the vision and language domains.

 {\bf Adversarial attack on graphs.}
Adversarial attack on graphs has recently been explored to show the robustness of  GNNs. \citet{dai2018adversarial} proposed different attacking strategies on both node and graph classification tasks. These strategies include  a greedy attack based on gradients (GradArgmax), a genetic algorithm-based attack (GeneticAlg), and a hierarchical Q-learning attack (RL-S2V).   \citet{zugner2018adversarial} introduced adversarial attacks on attributed graphs by following a greedy approximation scheme while making unnoticeable perturbations. \citet{zugner2019adversarial} used a meta-learning-based attack and showed a significant performance drop in classification accuracy even with small perturbations. In contrast to the above methods which attack node classifiers, \citet{bojchevski2019adversarial} attempted to attack unsupervised embedding algorithms like Deepwalk \cite{perozzi2014deepwalk} and node2vec \cite{grover2016node2vec} by  perturbing the adjacency matrix. All the aforementioned attacks make a common assumption that the attacker has access to the existing nodes in the graph; therefore, the attacker can perturb edges/features associated with those nodes. Gaining complete access to already existing nodes is a hard assumption and impractical. Therefore, there have been recent studies \citep{wang2018attack, sun2020adversarial, wang2020scalable}, which attack via {\em node injection}.

 {\bf Node injection attacks.}
Owing to the more realistic setting of attack, node injection attacks have gained popularity in the adversarial learning paradigm. Similar to RL-S2V, NIPA \citep{sun2020adversarial} used hierarchical Q-learning to generate labels and edges of the injected nodes. AFGSM \citep{wang2020scalable} provides an approximate closed-form solution for generating new features and edges of the injected nodes. AFGSM is a scalable method, which works well on both small and large-scale graphs and has shown performance at par, if not improved, against methods proposed in \cite{zugner2018adversarial, zugner2019adversarial} when changed according to node injection strategies. G-NIA \citep{tao2021single} introduces an attack with the extreme setting of a single-node injection using an optimization-based method and can even work with continuous attributed graphs, unlike previous methods. Later, TDGIA \citep{zou2021tdgia} utilizes a topological edge selection strategy and a smooth feature optimization objective to generate new edges and features of the injected nodes, respectively. Recently, there has been a study in node injection attacks which focuses on homophily preservation. \citet{fang2022gani} is a genetic algorithm based attack, trying to make unnoticeable perturbations in both structural and feature domains. The methods proposed by \citet{chen2022understanding} and \citet{tao2022adversarial} are developed as plug-ins on already existing graph injection attacks to improve their unnoticeability.
With respect to the time of perturbation, an adversarial attack can be classified into two types -- evasion and poison attacks. In an evasion attack \citep{dai2018adversarial}, the attack takes place on the test graph after the model has been trained. On the other hand, poison attack \citep{zugner2019adversarial, bojchevski2019adversarial} occurs on the training graph, and therefore, is comparatively  harder to solve because the graph learning protocol is compromised due to the attack. Both poison \citep{wang2020scalable, sun2020adversarial} and evasion \citep{tao2021single, zou2021tdgia, fang2022gani, chen2022understanding, tao2022adversarial} attacks have been studied for node injection.

Here we propose a node-injection-based poison attacking strategy that is capable of learning both discrete and continuous attributes through optimization. Unlike existing methods which mostly attack specific nodes, our attack is {\em class-specific}, attempting to misclassify nodes present in a given target class. A similar type of attack has been studied in the vision domain \citep{shafahi2018poison}.

\section{Preliminaries}

Let $G(V,A,X)$ be an undirected, unweighted and attributed network, where $A \in \{0,1\}^{N\times N}$ is the adjacency matrix of the graph and $X \in \{0,1\}^{N\times D}$ is the feature matrix such that $X^T=[x_1,x_2,\ldots, x_N]$, where $x_i \in \{0,1\}^{D}$ is the $D$-dimensional feature vector of $i^{th}$ node. 
Let $V$ be the set of nodes with $|V|$=$N$ and $V_L$ ($V_L\subset V$) be the set of labelled nodes where each node $u \in V_L$ is assigned a class $l_i$ such that $l_i \in L = \{l_1,l_2,\ldots,l_{|L|}\}$. In  semi-supervised node classification, the goal of a classifier $g : V \to L$ is to correctly classify the nodes present in $V\setminus V_L$. The objective of the node classification task can then be defined as:
\begin{align}
    \theta' = \arg\min_{\theta}\sum_{v \in V_L}  \mathcal{L}(g_{\theta}(G, v), l(v))\label{eq:mis-class}
\end{align}
Here $\theta$ is the set of parameters for the classifier $g$, $l(v)$ is the ground-truth class of node $v$, and $\mathcal{L}$ is the node classification loss (typically a cross-entropy function).

\begin{table}[!t]
    \centering
    \caption{Useful notations used throughout the paper.}
    \scalebox{0.92}{
    \begin{tabular}{c|p{55mm}}
    \hline
        {\bf Notation} & {\bf Denotation}  \\\hline\hline
        $G=(V,A,X)$ & Original graph \\
        
        $G'=(V',A',X')$ & Poisoned graph \\
        
        $L$ & Set of class labels \\
        
        $V$ & Set of original nodes \\
        
        $V_L$ & Set of labelled nodes \\
    
        $V_\A$ & Set of attacker nodes \\
        
        $V_b$ & Set of base class nodes ($b$ denotes the base class) \\
        
        $X'$ & Poisoned featured matrix \\
        
        $k$ & Number of injected attacker nodes \\
        
        $\Delta_e, \Delta_x$ & Budgets for edge and feature perturbations \\
        
        $Z_{\A}$ & Latent representation of attacker nodes \\
    
        $C_e, C_x$ & Candidate edges and features set \\
        
        $F_e, F_x$ & Multilayer Perceptrons \\
        
        $S_e, S_x$ & Scored edges and features \\
        \hline
    \end{tabular}}
    \label{tab:notation}
\end{table}

Table \ref{tab:notation} summarizes useful notations. We define a few important terminologies below:

\begin{definition} [Attacker Nodes]
    Given an original network $G$, a set of attacker nodes $V_\A$ are injected by an attacker $\A$ to poison the original network so as to deteriorate the performance of the downstream node classifier $g$.
\end{definition}

\begin{definition} [Target Class]
    Given the original network $G$ and a set of labels $L$, the target class $l_t\in L$ is a class whose associated nodes are targeted by an attacker $\A$. 
\end{definition}
  
\begin{definition} [Class-specific Poison Attack]
    Given an original network $G$, a target class $l_t$, attacker $\A$ injects nodes $V_\A$ along with their vicious features and edges to generate a poisoned network $G'$. The node classifier $g$ is trained on $G'$. The aim of the attacker is to let $g$ misclassify the nodes in the target class as much as possible.
\end{definition}

\section{Problem Formulation}\label{problem formulation}
In a network poison attack, we inject $V_\A$, a set of $k$ attacker nodes. We then obtain a poisoned graph $G'(V', A', X')$ with $V'=V \cup V_\A$, $A'=\begin{bmatrix}A & B\\ B^T & C\end{bmatrix}$ and $X'=\begin{bmatrix}X \\ X_{\A} \end{bmatrix}$. Here $B$ is a submatrix containing the links between nodes in $V_\A$ and $V$, and $C$ contains internal edges among nodes in $V_\A$. $X_{\A}^T = [x_{a_1},x_{a_2},\ldots,x_{a_k}]$ denotes the feature matrix of the attacker nodes. Note that we start with $B = 0$, $C = I$, and $X_\A = 0$ and use the obtained $G'$ as input to our framework pipeline. We iteratively update $G'$.

\begin{problem}[Node miscalssification]
\label{Node miscalssification problem}
    The goal is to learn $B$, $C$ and $X_{\A}$ such that $g$ misclassifies nodes in $V_t\subset V$ having class $l_t$. The objective function of our poison attack can be defined as a bi-level optimization problem as follows:
    \begin{flalign}
        &\max_{B, C, X_{\A}} \sum_{v \in V_t} \mathbb{I}(g_{\theta'}(G', v) \neq  l_t) \label{eq:obj}\\
        &:\|A'\|_0 - \|A\|_0 \leq 2\Delta_e, \|X'\|_0 - \|X\|_0 \leq \Delta_x \label{eq:cond2}
    \end{flalign}
\end{problem}
\noindent Here $\mathbb{I}$ is the indicator function. 
Equation \ref{eq:obj} maximizes the number of misclassified nodes belonging to the target class. $\theta'$ are the optimal parameters we receive from Equation \ref{eq:mis-class}, by tuning the classifier $g$ on the poisoned graph $G'$. Equation \ref{eq:cond2} provides an upper bound to the total number of updates in $B$, $C$ and $X_{\A}$. Note that we do not modify existing adjacency matrix $A$ and features $X$. Since ours is a poisoning attack, the evaluation of the model is conducted after training $g$ on the poisoned graph.

The goal of a poison attack is to infect the training set, thereby making an assumption that all the attacker nodes in $V_\A$ also belong to the training set. 
However, an attacker does not have control over labelling, which poses a major challenge. If not accounted for, the attacker nodes have a high chance of being detected as an outlier and consequently, removed from the network. We address this challenge by making the adversaries belong to a benign class.

\begin{problem}[Hiding attacker nodes]
\label{Hiding problem}
The goal is to camouflage the attacker nodes $V_\A$ into a class $l_b$ which we call our ``base class''. The objective can be defined as follows:
\begin{flalign}\label{eq:hid_node}
    \arg \min_\lambda Im_{\lambda}(V_\A,V_b)
\end{flalign}
Here $V_{b}$ denotes the set of nodes in the base class. $Im_{\lambda}$ measures the visual distance between $V_\A$ and $V_b$ with $\lambda$ denoting a set of parameters. Here visual distance is used to quantify the closeness of both feature and topological properties of the given two sets of nodes.
\end{problem}
The above formulation restricts the nodes in $V_\A$ to look similar to those in $V_b$. This would help mislead the annotator in labelling the attacker nodes as the base class.

\section{Proposed Framework}
\begin{figure*}
  \includegraphics[width=\textwidth]{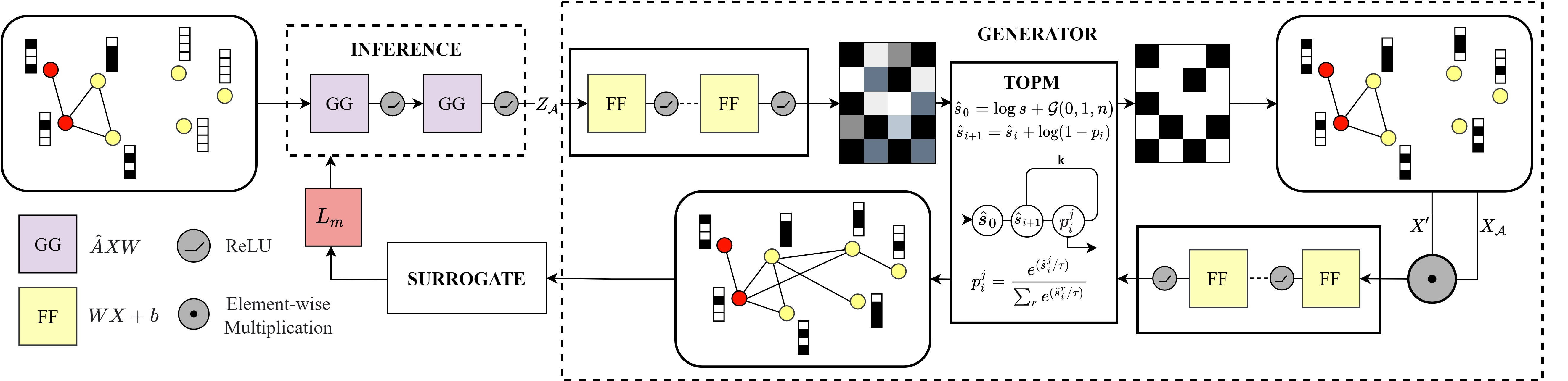}
  \caption{
  A Schematic diagram of \model. The two modules, {\bf Inference} and {\bf Generator}, run in series to produce the poisoned graph $G'(V', A', X')$. The {\bf Surrogate} is trained at every iteration to solve the bi-level optimization, and the loss thus obtained is minimized using gradient descent.
}\label{fig: model}
\end{figure*}

Here we describe our proposed attack framework, \model\ ({\bf N}ode {\bf I}njection for {\bf C}lass-specific Networ{\bf K} Po{\bf I}soning). 
The attack takes place in a poison setting, i.e., the node classifier would misclassify nodes in the target class on the poisoned graph. The proposed model, as shown in Figure \ref{fig: model}, is a two-part network -- (i) inference model, and (ii) generator model. 
The inference model is a graph encoder which learns a latent representation for each attacker node injected in the graph. The generator model comprises two modules that run sequentially  to construct the poisoned graph -- (i) a scoring module which scores each edge and feature such that their combination results in misclassification, and (ii) a top-$m$ selector which chooses the best edges/features from scores generated by the scoring module. 

The top-$m$ module is used to ensure a differentiable method to impose two constraints on the number of perturbations -- $\Delta_e$ and $\Delta_x$, which are maximum number of permissible changes in edges and features, respectively. 

Even though we impose a budget constraint, our injected attacker nodes could be exposed and subsequently be removed from the graph. Therefore, we also propose sophisticated hiding techniques, which camouflage's attacker nodes with the nodes of a chosen base class. Once we have the attacker nodes labelled as $l_b$, we leverage it to misclassify our target nodes into $l_b$. 
We, therefore, update Equation \ref{eq:obj} as: 
\begin{flalign}
    &\max_{B, C, X_{\A}} \sum_{v \in V_t} \mathbb{I}(g_{\theta'}(G', v) = l_b) \label{eq:obj_ours}
\end{flalign}
Post hiding, we use Equation \ref{eq:loss-mis} to bring close the embeddings of attacker nodes and base class nodes so that when the classifier is retrained on the poisoned graph, the decision boundary in the embedding space would be expected to rotate such that the attacker nodes are labelled as base class. The target nodes being close to the attacker nodes would also be included in the base class, resulting in misclassification.

\subsection{Inference Model}\label{inference-model}
For our implementation, we introduce an intermediate graph $G_P(V',A_P,X_P)$, with $A_P \in \{0,1\}^{(N+k)\times (N+k)}$ and $X_P \in \{0,1\}^{(N+k)\times D}$ defined as:
\begin{align}\label{eq:Ap}
    (A_P)_{i,j} = 
    \begin{cases}
    A'_{i,j}, \;\;\; & 0\leq i,j<N\\
    1,  & \text{otherwise}
    \end{cases}
\end{align}
\begin{align}\label{eq:Xp}
    (X_P)_{i,j} = 
    \begin{cases}
    X'_{i,j}, \;\;\; & 0\leq i<N,0\leq j<D\\
    1,  & \text{otherwise}
    \end{cases}
\end{align}
We use $A_P$ and $X_P$ as two inputs to the inference model. 

We use Graph Autoencoders (GAEs) \cite{kipf2016variational} to learn a joint representation of $A_P$ and $X_P$ given by the latent variable $Z$ which is parameterized by a two-layer Graph Convolutional Network (GCN) as follows:
\begin{flalign}\label{eq:infer}
    &Z = \sigma_2(\hat{A}_P \sigma_1 (\hat{A}_P X_PW^{(0)})W^{(1)})
\end{flalign}
Here $ \hat{A}_P$ represents the normalized adjacency matrix, described as $\hat{A}_P = \hat{D}^{-\frac{1}{2}}(A_P+I_n) \hat{D}^{-\frac{1}{2}}$, $\hat{D}$ is a diagonal matrix with $\hat{D}_{ii}=\sum_j(A_P+I_n)_{ij}$, and $I_n$ is the $n${th}-order identity matrix. $W^{(l)}$ denote the $l${th}-layer weights of the GCN. $\sigma_1$ and $\sigma_2$ are activation functions.

The inference model is essentially a GCN which propagates knowledge through edges. Therefore, making all the candidate edges and features as $1s$ in $A_P$ and $X_P$, respectively (Equations \ref{eq:Ap} and \ref{eq:Xp}) allows the attacker nodes to indirectly have access to the entire graph during encoding. This eventually results in more meaningful and useful latent embeddings for the attacker nodes.

\subsection{Generator Model}\label{generator}
Once we encode $A_P$ and $X_P$ into $Z$, our objective is to use the latent representation of the attacker nodes $Z_\A$ to generate both feature and edge perturbations. This is done by using two modules -- {\em scoring} and {\em top-$m$ selector}. The scoring module generates scores for each edge and feature in the candidate set, defined as $C_e$ and $C_x$, respectively. The top-$m$ module enforces the budget with a differentiable method and selects the edges and features with the highest scores.

\subsubsection{Scoring Module}\label{scoring}
We derive our set of candidate edges from the submatrices $B$ and $C_U$ in $A'$, where $C_U$ is the upper triangle of $C$ not including the diagonal elements. We define ($B_{ij} = 0$ or $C_{ij} = 0) \implies e_{ij} \in C_e$, where $e_{ij}$ denotes an edge between nodes $i$ and $j$. 
On the other hand, all the attributes present in $X_\A$ constitute our candidate features. Therefore, $x_{ij} \in X_\A \iff x_{ij} \in C_x$, where $x_{ij}$ denotes the $j^{th}$ feature entry for node $i$. The scoring of features and edges is performed in series as described below:

$Z_\A$ is fed to a Multilayer Perceptron (MLP) $F_x$ as:
\begin{align}
    S_x = F_x(Z_\A),\text{ where } Z_\A, S_x \in \mathbb{R}^{k \times D} 
\end{align}
Here $k$ is the number of attacker nodes, and $F_x$ is a one-to-one mapping with $S_x$ representing the score for each candidate feature. $S_x$ is then passed through the top-$m$ selector, which selects $\Delta_x$ candidate values to be switched to 1s. The following equation represents the same:
\begin{align}\label{eq:topk-feature}
    X_\A = TOPM_{\Delta_x}(S_x), \text {where } X_\A \in \mathbb{R}^{k \times D} 
\end{align}
where $TOPM_{\Delta_x}(S_x)$ denotes top-$m$ selector for selecting the top $\Delta_x$ features according to their scores.  
Once $X_\A$ is generated, we use it to score the candidate edges as follows:
\begin{align}
    S_e &= F_e(X' \odot X_\A), \text{ where } S_e \in \mathbb{R}^{(k+N) \times k} \label{eq:edge-scoring}\\
    S_e &= S_e - \{s_e\} \text{ } \forall e \notin C_e 
\end{align}
where $X'=\begin{bmatrix}X \\ X_\A\end{bmatrix}$, $\odot$ denotes Vector-wise Dot-product for each pair of vectors, 
$s_e$ denotes the obtained score for an edge $e$, and $F_e$ is an MLP. Equation \ref{eq:edge-scoring} is used to score all candidate edges by element-wise multiplying the features of the pair of nodes connecting each edge. Post scoring, top-$m$ selector is used to select $\Delta_e$ edges as follows:
\begin{align}
    T_e = TOPM_{\Delta_e}(S_e), \text{ where } T_e \in \mathbb{R}^{k(k/2+N) \times 1} \nonumber
\end{align}
$T_e$ is further reshaped to matrix form $A'$ by concatenating with original adjacency matrix $A$.

We use a GCN-based model to evaluate our attack's performance. We call this our surrogate model which aims to mimic the original classifier $g$.

\subsubsection{Top-$m$ Selector}
Given a vector $s = [s_1,s_2,s_3, \ldots,s_n]$ of $n$ scores, the goal of top-$m$ selector is to provide a differentiable method, which samples a binary vector $t$ of size $n$ with exactly $m$ $1$s denoting the $m$-highest scores in $s$. 

We adapt Relaxed Subset Sampling \cite{xie2019reparameterizable} for our use case. The algorithm aims to iteratively select the $i${th} maximum element in the $i${th} iteration. In every iteration, first the score vector is updated as:
\begin{flalign}
    \hat{s}_{i+1} = \hat{s}_{i} + \log (1-p_i);
    \text{ with } \hat{s}_0 = \log s + Gumbel(0,1,n) \nonumber
\end{flalign}
where $Gumbel(0,1,n)$ is an $n$-dimensional vector sampled from the gumbel distribution. The maximum score is then selected from the updated vector using $tempered$ $softmax$ to obtain a probability distribution $p_i = \{p_i^1,p_i^2,...p_i^n\}$ where,
    $p_i^j = \frac{e^{(\hat{s}_i^j/\tau)}}{\sum_r e^{(\hat{s}_i^r/\tau)}}$.
Here $p_i^j$ denotes the probability of the $j${th} element of our input vector $s$ being the $i${th} maximum element. $\tau$ is the temperature parameter. After $m$ iterations, we obtain our resultant binary vector $t = \sum_{i=1}^m p_i$. Algorithm \ref{algo:subset_sample} shows the pseudo-code of the Top-$m$ module.

\begin{algorithm}[!t]
\SetAlgoLined
\SetKwInOut{Input}{Input}
\SetKwInOut{Output}{Output}
\Input{$s=[s_1, s_2, s_3...s_n],m,\tau$}
\Output{Relaxed $k$-hot vector $t=[t_1,t_2,t_3,...t_n]$}
    $\hat{s} = \log s+Gumbel(0,1,n)$\\
    $p = [0,0,...0](n-times)$\\
    $t = [0,0,...0](n-times)$\\
    \For{$i\gets1$ \KwTo m}{
        $\hat{s}=\hat{s}+log(1-p)$\\
        $p=softmax(\hat{s}/\tau)$\\
        $t=t+p$
    }
\BlankLine
 \caption{TOP-$m$ Selector}
 \label{algo:subset_sample}
\end{algorithm}

\subsection{Hiding Attackers}\label{hiding-framework}
\begin{figure}[!t]
  \includegraphics[width=\columnwidth]{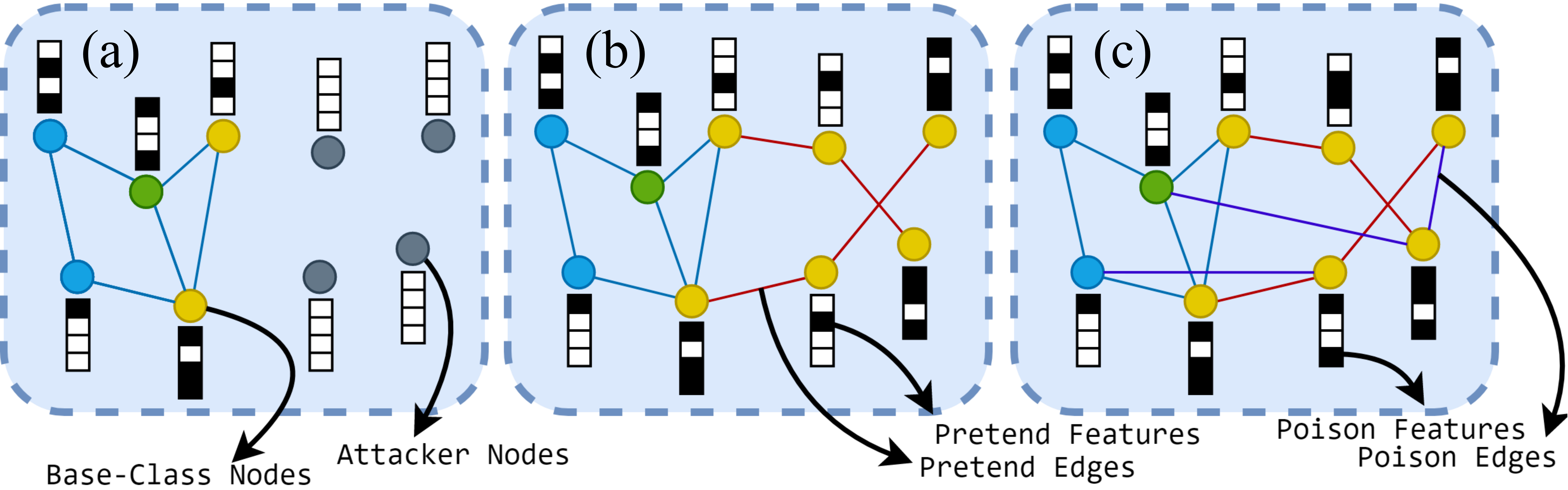}
\caption{(a) Clean graph just after node injection. (b) Resultant graph post hiding injected attacker nodes. The attacker nodes pretend to be in base class (yellow) using pretend edges and features. (c) Graph after attack where new poison edges are introduced along with pretend edges. Note that the attributes of attacker nodes have changed a little owing to the regularization loss $L_f$.}
  \label{fig:graph}
\end{figure}

We define a hiding module to solve Problem \ref{Hiding problem} (Section \ref{problem formulation}) by camouflaging the attacker nodes into a base class $l_b$, which allows them to be both included and labelled during the training process. We achieve the same by initialising the nodes in $V_{\A}$ with some \textit{pretend edges} and {\em features}, assuring that they belong to the base class $l_b$. Our hiding framework is applied before the attack framework (Figure \ref{fig:graph}). Therefore, it can be understood as a pre-processing strategy. 

\subsubsection{Pretend Edges}
Our goal is to make any node $i \in V_\A$ look like it belongs to $V_b$. To achieve this, we sample $d_i$, the degree of node $i$ from the power-law degree distribution of nodes in $V_b$, which we obtain by fitting their Complementary Cumulative Distribution Function (CCDF). If $i \in V_\A$, we add a certain fraction of $d_i$ edges from $i$ to nodes in $V_b$. The connections are made based on the Barabási–Albert model \citep{barabasi1999emergence}, where the probability of connection is directly proportional to the degree of nodes in $V_b$. We repeat the process for every node $i \in V_\A$ and add all such pairs $(i,j)$, where $j \in V_b$, to our final set of pretend edges, $PE$.
Note that we sample $d_i$ such that $d_i>d_{\mu}$, where $d_{\mu}$ is the average degree of the input graph. This is to ensure that we have enough edges to connect with the nodes in $V_b$.

\subsubsection{Pretend Features}\label{sec:pretend_feature}
For pretend features, our objective is to initialise all $u \in V_{\A}$ with features $x_u \in X_{P}$ such that $u$ pretends to belong to $V_{b}$.
All nodes belonging to the same class ought to have some feature commonalities that differentiate them from the nodes in other classes. The idea is to fit the features of nodes in $l_b$ class to a distribution $P_{b}(X)$, from which we can sample $X_u, \; \forall u \in V_\A$. 
We achieve the same by using the Conditional VAE (CVAE) \citep{sohn2015learning}) architecture. 
In contrast to VAEs, CVAEs have their output conditioned on $l_i$-class label which prevents repeating the training process for different base classes. The CVAE objective can be formulated as,
\begin{align}
    L(\phi, \theta, X, l_i) &= − KL(q_{\phi}(z|X, l_i)\parallel p_{\theta}(z|l_i))\nonumber\\  &+  E_{q_{\phi}}(z|X,l_i)[logp_{\theta}(X|z,l_i)]
\end{align}
During testing or generative phase, we give our desired base class $l_b$ as $l_i$ and use the decoder to generate $x_{u}$, $\forall u \in V_{\A}$. We stack all such $x_u's$ to form our \textit{pretend features} matrix $PF$ such that $PF \in \{0,1\}^{k\times D}$. We sample the features for each attacker node independently.
Once we obtain our \textit{pretend edges} $PE$ and {\em pretend features} $PF$, we replace and re-initialise the input to our attack pipeline, $A'$ and $X'$, as,
\begin{align}\label{A' update}
    A'_{i,j} = 
    \begin{cases}
    A'_{i,j} \text{ ,} \;\;\; & 0\leq i,j<N\\
    1 \text{ ,}  & (i,j) \;\; or \;\;(j,i) \in PE \\
    0 \text{ ,} & otherwise
    \end{cases}
\end{align}
\begin{align}\label{X' update}
    X'_{i,j} = 
    \begin{cases}
    X'_{i,j} \text{ ,} \;\;\; & 0\leq i<N,0\leq j<D\\
    PF_{(i-N),j} \text{ ,}  & otherwise
    \end{cases}
\end{align}

We use these matrices, $A'$ and $X'$, as inputs to our inference model, instead of what is defined in Section \ref{problem formulation}. Note that according to the definition of $C_e$ in Section \ref{scoring}, this also changes our candidate edge set as the obtained \textit{pretend edges} are to be masked from the attack pipeline as well.

Algorithm \ref{algo:attack_model} shows the pseudo-code for \model.

\begin{algorithm}[!t]
\SetKwInOut{Input}{Input}
\SetKwInOut{Output}{Output}
\Input{Clean graph $G(A, X)$, label set $L$, budgets $\Delta_e, \Delta_x \A$, base class $l_b$, target class $l_t$, pre labeled set $V_L$, HIDE}
\Output{Poisoned graph $G(A', X')$, and updated labeled set $V'_L$}
\If{HIDE}
{Connect nodes in $V_\A$ with base class nodes to hide.\\ Assign features to each node $i \in V_\A$ from the distribution of features in base class.\\ Update $A$ and $X$ with attacker nodes.}
\Repeat{iterations $<$ Max iterations}
{Compute embeddings of $A,X$ using Equation \ref{eq:infer}.\\
Calculate scores for each candidate features as
$S_x \leftarrow F_x(Z_\A)$\\
Select top scorers with a constraint on budget using TOP-m Selector as described in Equation \ref{eq:topk-feature}.\\
Update the feature matrix as
$X' \leftarrow [X^T, X_\A^T]^T$\\
Use updated features to generate candidate edges as
$S_e \leftarrow F_e(X' \odot X')$\\
Pass scores through TOPM module to get the updated adjacency matrix as
$A' \leftarrow reshape(TOPM_{\Delta_e}(S_e))$\\
Train the surrogate model from the updated Adjacency and feature matrix
$L_m \leftarrow Surrogate(A', X')$\\
\If{HIDE}{Calculate change in feature using Equation \ref{eq:loss-feat}}}
 \caption{Attack Model}
 \label{algo:attack_model}
\end{algorithm}

\subsection{Loss Function}\label{loss}

We use average cross-entropy for misclassification and hiding feature loss. For misclassification loss ($L_m$), we compare embeddings of attacker and mean target class nodes as:
\begin{align}\label{eq:loss-mis}
    L_m = \sum_{i=1}^{k} \frac{-\sum_{j=1}^{D} Q_{\mu tj} \log Q_{ij}}{k}
\end{align}
where $Q_{\mu t}$ is the mean embedding (surrogate model) of the nodes belonging to target class $l_t$, and $Q_i$ is $i$th attacker node's embedding (surrogate). While in hiding feature loss ($L_f$), we minimize the distance between matrices $PF$ and $X_\A$ using cross-entropy as follows:
\begin{align}\label{eq:loss-feat}
    L_f = -\sum_{i=1}^{k} \frac{\bar{X}_i\log\bar{PF}_{i}}{k}
\end{align}
where $\bar{X}_i$ is the $i$th normalized feature vector of $X_\A$ we obtain from the generator module, and $\bar{PF}_{i}$ is the corresponding normalized \textit{pretend} feature vector.

We train our adversarial attack network by minimising the following loss function L,
\begin{align}\label{total_loss}
    L = L_m + \alpha L_f
\end{align}
Note that $\alpha \neq 0$ when we employ our hiding framework.

\subsection{Time Complexity}\label{app:timecomplexity}

The model architecture constitutes  GCNs, MLPs and top-$m$ modules with some time consumed by pre-processing in the hiding module. (i) For a single-layer GCN, we need to compute $Z = f(X,A) = \sigma(\hat{A}XW^{(0)})$.
If $\hat{A} \in \mathbb{R}^{N \times N}$, $X \in \mathbb{R}^{N \times D}$ and $W^{(0)} \in \mathbb{R}^{D \times H}$, then $Z \in \mathbb{R}^{N \times H}$. The time complexity of the matrix multiplication before the activation function $\sigma$ can be computed in $\mathcal{O}({E}D+NDH)$, where ${E}$ is the number of edges in matrix $A$. The time complexity of executing an activation function is linear in the size of input.
  Given an input $x \in \mathbb{R}^{m_1 \times m_2}$, a hidden layer $h \in \mathbb{R}^{m_2 \times m_3}$, and an output layer $o \in \mathbb{R}^{m_3}$, the complexity for a forward pass in an MLP is simply the time to compute the matrix multiplication, i.e., $\mathcal{O}(m_1m_2m_3)$, with activation function executing in linear time on the input. The time complexity of the top-$m$ selector is $m$ times the size of input, as it executes tempered softmax at each iteration which runs in linear time. So given an input $x \in \mathbb{R}^n$, the time complexity would be $\mathcal{O}(mn)$.
(ii) The complexity of a 2-layer GCN in the inference model will take $\mathcal{O}(2({E}D+ND^2))$. There are two MLPs -- $F_x$ and $F_e$. The time complexity of $F_x$ and $F_e$ would be  $\mathcal{O}(3k D^2+2k D)$ and $\mathcal{O}(5Nk D^2+ 6Nk D)$, respectively. The time complexity of $2$ top-$m$ selectors, $TOPM_{\Delta_x}$ and $TOPM_{\Delta_e}$ would be $\mathcal{O}(\Delta_x k D)$ and $\mathcal{O}(\Delta_e k (N+k))$, respectively.

Therefore, the overall complexity of \model\ is $\mathcal{O}(D({E}+k (N D+ \Delta_x)) +\Delta_e k (N+k))$.

\section{Experiments}

\subsection{Datasets}\label{app:dataset} 
We use four widely-used benchmark datasets for our experiments. 

$\blacksquare$ {\bf Cora} \citep{mccallum2000automating} is a citation network wherein nodes are scientific papers, and  edges indicate citation relationships among them. Each node has a  bag-of-word attribute vector obtained from processing the text in the corresponding document. The set of publications is divided into $7$ different classes.

$\blacksquare$ {\bf Citeseer} \citep{10.1145/276675.276685}
is also a citation network, similar to Cora. Here papers are divided into $6$ classes.

$\blacksquare$ {\bf PolBlogs} \citep{adamic2005political}
is a network of political weblogs in the US with nodes representing blogs and edges referring to hyperlinks among blogs. Each blog has an associated political orientation, representing its class. There are two classes - Democratic and Republican.

$\blacksquare$ The {\bf Reddit} \citep{hamilton2017inductive} network contains posts represented by nodes and edges showing post-to-post relationships. The nodes are divided into $41$ different classes representing the subreddit the corresponding posts are part of. The attribute vector is obtained from the GloVe \citep{pennington2014glove} word embedding on the title and comments. Table \ref{tab:dataset} shows the statistics of the datasets.

We train GCN for node classification before and after graph poisoning to study the performance of  \model. The citation networks have discrete attributes; therefore, showing our performance on them validates the working of our top-$m$ selector module. We convert them into undirected networks. Polblogs is an attribute-less network, thus providing the efficacy of \model's malicious edge generation. The Reddit dataset has continuous attributes, which shows \model's adaptability to work on continuous attribute space. It also contains large number of nodes, which checks our model's scalability. 

\begin{table}[!t]
    \centering
    \caption{Network statistics. PolBlog is a featureless graph.}
    \label{tab:dataset}
    \scalebox{1}{
    \begin{tabular}{c|c|c|c|c|c}
        \hline
        Dataset & $N$ & $|E|$ & $d_{avg}$ & $D$ & $|L|$\\\hline
        Cora & 2708 & 6632 & 4.898 & 1433 & 7 \\
        Citeseer & 2110 & 3694 & 3.501 & 3703 & 6 \\
        PolBlogs & 1086 & 9502 & 17.499 & - & 2 \\
        Reddit & 10004 & 31754 & 6.348 & 602 & 41 \\
        \hline
    \end{tabular}}
\end{table}
\subsection{Baseline Methods}\label{app:details_baseline}
\model\ is a node injection based poison attack and hence, we choose baseline attacking strategies based on two conditions -- (i) poison, and (ii) node injection. We compare \model\ against two heuristic and two recent methods.

$\blacksquare$ {\bf Random.}
This attack follows the Erdos-Rényi model \citep{erdos1960evolution} for generating new edges. Each new edge is selected with probability $p = ||A||_0/|V|^2$, which is the normalized total degree of $G$. The process continues until the budget is exhausted. For feature generation, attributes of existing nodes are averaged, and the resultant vector defines the attribute of the attacker nodes.

$\blacksquare$ {\bf Preferential.}
The attack is based on Barabási–Albert model \citep{barabasi1999emergence}. Nodes are injected sequentially, and edges connecting the new node to the rest are selected through preferential attachment. We follow a feature generation procedure identical to that of a random attack.

$\blacksquare$ {\bf NIPA.} This
 is a node injection poison attack, which employs hierarchical Q-learning to generate adversarial edges \citep{sun2020adversarial}. The adjacency matrix learnt by NIPA is used as it is. For attribute generation, it takes an average over the feature vectors of pre-existing nodes and adds some Gaussian noise to it.  NIPA is a global attack, and therefore, we modify the test set to contain nodes only belonging to the target class.
 
$\blacksquare$ {\bf AFGSM.} This
 is a greedy attack, which uses an approximate closed-form solution for generating edges and attributes making it scalable w.r.t the size of the network \citep{wang2020scalable}. Since AFGSM is a single-node target attack, we modify it to suit our setting by measuring scores corresponding to each target node. We then sum the obtained scores corresponding to all the target nodes followed by adversarial selection. Therefore, the perturbation which has a higher total score is more likely to be chosen depending on the budget. We train the adaptive variant of AFGSM since it trains the surrogate dynamically during training, thus mimicking a poison attack.

\begin{table}[!th]
    \centering
    \caption{Accuracy of GCN-based node classification  after adversarial attack, across varying budgets. $r$ controls the budget (Section \ref{sec:setup}).  {\color{red}Red} ({\color{blue}blue}) color represents the best (second ranked) model. Lower value indicates better attack. Values within $[.]$ in the first column indicate the classification accuracy on clean graph. See Table \ref{tab:gat} for the same using GAT and GraphSAGE as node classifiers.}
    \scalebox{0.8}{
    \begin{tabular}{c|c|cccc}
    \hline
        {\bf Dataset} & {\bf Method} & {\bf $r=0.03$} & {\bf $r=0.07$}& {\bf $r=0.10$}& {\bf $r=0.15$}\\\hline\hline
        \multirow{6}{*}{\makecell{ Cora \\ $[0.8189]$}} & Random & 0.7911 & 0.8050 & 0.8078 & 0.7994 \\
        & Preferential & 0.7967 & 0.7911 & 0.8106 & 0.7939 \\
        & NIPA & 0.7632 & 0.7465 & 0.7716 & 0.7994 \\
        & AFGSM & 0.8161 & 0.7855 & 0.7827 & 0.7437 \\
        & \model & {\color{red}0.6741} & {\color{red}0.5070} & {\color{red}0.4345} & {\color{red}0.3983} \\
        & \model\ (hide) & {\color{blue}0.7409} & {\color{blue}0.7270} & {\color{blue}0.7103} & {\color{blue}0.7214} \\
        \hline
        \multirow{6}{*}{\makecell{ Citeseer \\ $[0.7395]$}} & Random & 0.7474 & 0.7500 & 0.7553 & {\color{blue}0.7316} \\
        & Preferential & 0.7526 & 0.7605 & 0.7579 & 0.7342 \\
        & NIPA & 0.7500 & 0.7474 & 0.7447 & 0.7579 \\
        & AFGSM & 0.7421 & 0.7947 & 0.7421 & 0.7342 \\
        & \model & {\color{red}0.6632} & {\color{red}0.5632} & {\color{red}0.5500} & {\color{red}0.4526} \\
        & \model\ (hide) & {\color{blue}0.7131} & {\color{blue}0.7263} & {\color{blue}0.7316} & {\color{blue}0.7316} \\
        \hline
        \multirow{6}{*}{\makecell{ Polblogs \\ $[0.9316]$}} & Random & 0.9203 & 0.9227 & 0.9034 & 0.8913 \\
        & Preferential & 0.9251 & 0.9275 & 0.9203 & 0.9300 \\
        & NIPA & {\color{blue}0.9058} & {\color{blue}0.8986} & {\color{blue}0.8551} & 0.8937 \\
        & AFGSM & 0.9420 & 0.9420 & 0.9203 & 0.9565 \\
        & \model & {\color{red}0.8164} & {\color{red}0.7657} & {\color{red}0.7657} & {\color{red}0.6522}\\
        & \model\ (hide) & 0.9109 & 0.9082 & 0.9034 & {\color{blue}0.8527}\\
        \hline
        \multirow{6}{*}{\makecell{ Reddit \\ $[0.9348]$}} & Random & 0.9144 & 0.9125 & 0.9163 & 0.9240 \\
        & Preferential & 0.9183 & 0.9144 & 0.9144 & 0.9221 \\
        & NIPA & 0.9144 & 0.9240 & 0.9202 & 0.9259 \\
        & AFGSM & {\color{red}0.8821} & {\color{blue}0.8460} & {\color{blue}0.8441} & {\color{blue}0.8536} \\
        & \model & {\color{blue}0.8954} & {\color{red}0.7529} & {\color{red}0.8213} & {\color{red}0.7795} \\
        & \model\ (hide) & 0.9049 & 0.8935 & 0.9182 & 0.8631 \\
        \hline
    \end{tabular}}
    \label{tab:class-results}
\end{table}

\subsection{Experimental Setup}\label{sec:setup}

To measure the performance of our model, we use the poisoned graph to obtain the classification accuracy w.r.t nodes in the target class, and compare the same with that of the clean graph. The lower the accuracy, the better the attack. Note that we report the accuracy of the node classifier w.r.t only the $l_t$-labelled nodes in the test set. We inject $k = r|V_t|$ nodes, where the controlling parameter $r$ denotes the ratio of injected nodes to target class nodes.

Since ours is a poison attack, we evaluate the accuracy after training our surrogate on the poison graph. We split each dataset into $10\%, 10\%, 80\%$ of the total nodes for training, validation, and testing, respectively, and use a two-layer GCN with a structure same as the one described in Section \ref{inference-model} as our surrogate model. Following the definition of poison attack, we include our attacker nodes $V_\A$ into the training set.

\begin{table*}[!t]
    \centering
    \caption{We repeat the experiment shown in Table \ref{tab:class-results}, but with GAT and GraphSAGE as our node classifiers. We do not show the results of random and preferential models as they are the worst among all.  Values within $[x,y]$ in the first column indicate the classification accuracy on the clean graphs obtained from GAT ($x$) and GraphSAGE ($y$).}\label{tab:gat}
    \scalebox{0.65}{
    \begin{tabular}{c|c|cccc|cccc}
    \hline
    & & \multicolumn{4}{c}{GAT} & \multicolumn{4}{c}{GraphSAGE} \\\cline{3-10}
        {\bf Dataset} & {\bf Method} & {\bf $r=0.03$} & {\bf $r=0.07$}& {\bf $r=0.10$}& {\bf $r=0.15$} & {\bf $r=0.03$} & {\bf $r=0.07$}& {\bf $r=0.10$}& {\bf $r=0.15$}\\\hline\hline
        \multirow{6}{*}{\makecell{ Cora \\ $[0.8134, 0.8078]$}} & NIPA & 0.7961 & 0.7701 & 0.7725 & 0.8144 & {\color{blue}0.7658} & 0.8006 & 0.7697 & 0.7950 \\
        & AFGSM & 0.8056 & {\color{blue}0.7313} & 0.8324 & 0.7452 & 0.7750 & {\color{blue}0.7486} & 0.8464 & 0.7928 \\
        & \model & {\color{red}0.6500} & {\color{red}0.6096} & {\color{red}0.4266} & {\color{red}0.5311} & 0.7806 & {\color{red}0.6994} & {\color{red}0.6441} & {\color{red}0.6130} \\
        & \model\ (hide) & {\color{blue}0.7000} & 0.7331 & {\color{blue}0.7345} & {\color{blue}0.7288} & {\color{red}0.7556}  & 0.7640 & {\color{blue}0.7006} & {\color{blue}0.7853} \\
        \hline
        \multirow{6}{*}{\makecell{ Citeseer \\ $[0.8000, 0.7947]$}} & NIPA & 0.8449 & 0.7506 & 0.7747 & 0.7755 & 0.7834 & 0.7224 & 0.7696 & 0.7807\\
        & AFGSM & 0.8080 & {\color{blue}0.7242} & {\color{blue}0.7519} & 0.7727 & {\color{red}0.7314} & {\color{blue}0.7010} & {\color{blue}0.7114} & {\color{blue}0.7424}\\
        & \model & {\color{red}0.7325} & {\color{red}0.5349} & {\color{red}0.6211} & {\color{red}0.3368} & {\color{blue}0.7377} & {\color{red}0.6925} & {\color{red}0.7010} & {\color{red}0.6247} \\
        & \model\ (hide) & {\color{blue}0.7844} & 0.8062 & 0.7655 & {\color{blue}0.7558} & 0.7766 & 0.7726 & 0.7758 & 0.7635 \\
        \hline
        \multirow{6}{*}{\makecell{ Polblogs \\ $[0.9444, 0.9444]$}} & NIPA & 0.8846 & 0.9502 & 0.8744 & {\color{blue}0.9329} & {\color{blue}0.8918} & 0.9005 & {\color{blue}0.8814} & {\color{blue}0.8472}\\
        & AFGSM & 0.9231 & {\color{blue}0.9151} & 0.9372 & 0.9425 & 0.9591 & 0.9599 & 0.9721 & 0.9264\\
        & \model & {\color{red}0.8155} & {\color{red}0.8019} & {\color{blue}0.8406} & {\color{red}0.7745} & {\color{red}0.8762} & {\color{red}0.7971} & {\color{red}0.8092} & {\color{red}0.7623} \\
        & \model\ (hide) & {\color{blue}0.8835} & 0.9469 & {\color{red}0.8188} & 0.9461 & 0.9417 & {\color{blue}0.8575} & 0.9058 & 0.9069 \\
        \hline
        \multirow{6}{*}{\makecell{ Reddit \\ $[0.9087, 0.9563]$}} & NIPA & 0.9032 & 0.8960 & 0.8807 & 0.8639 & 0.9279 & 0.9357 & 0.9545 & 0.9565\\
        & AFGSM & {\color{red}0.8539} & {\color{blue}0.8623} & {\color{blue}0.8617} & 0.8480 & 0.9336 & 0.9302 & 0.9034 & 0.9212\\
        & \model & {\color{blue}0.8740} & {\color{red}0.7699} & {\color{red}0.7211} & {\color{red}0.7324} & {\color{red}0.8836} & {\color{red}0.7965} & {\color{red}0.7381} & {\color{red}0.7154}\\
        & \model\ (hide) & 0.9103 & 0.8918 & 0.8710 & {\color{blue}0.8368} & {\color{blue}0.9103} & {\color{blue}0.9013} & {\color{blue}0.7780} & {\color{blue}0.8273} \\
        \hline
    \end{tabular}}
    \label{tab:transfer-attack}
\end{table*}

We trained our models for 50 epochs and used the learnt graph with minimum classification accuracy as our attack graph. There are 6 layers in $F_e$ with $5D/4$, $D$, $D/2$, $D/2$, $D/2$, and $D/16$ being the respective number of hidden nodes in layers. In $F_x$, we have 3 layers with number of hidden nodes being $5D/4$ and $5D/4$. For \model\ (hide), we used $0.7$ as parameter for internal degree. $\alpha$ is set to $0.1$. We use Adaptive Moment Estimation (Adam) as the gradient descent optimizer. Our code is written in PyTorch. Due to the unavailability of NIPA's original source code, we used DeepRobust \citep{li2020deeprobust} library's implementation.

{\bf Budget.} We use two budgets $\Delta_e$ and $\Delta_x$ for constraining edge and feature perturbations, respectively. Formally, we set $\Delta_e = kd_{avg}$, where $d_{avg}$ denotes the average degree of nodes in the graph. Such a formulation helps in retaining the average degree of nodes in the clean graph to that in the  poisoned graph, which is more indicative of a real-world attack. For discrete datasets, we set $\Delta_x = \frac{\sum_{i=1}^{N}\sum_{j=1}^{D}X_{ij}}{N}$, where $X_{ij}$ denotes the $j$th attribute of node $i$. For continuous datasets like Reddit, we limit feature perturbations by enforcing the original features' range over them. We do this by replacing $TOPM_{\Delta_x}$ from Equation \ref{eq:topk-feature} with a sigmoid function followed by mapping the values from $[0,1]$ to the target range. PolBlogs, being a featureless graph, does not require feature generation. Therefore, we directly use the output of our inference model $Z$ to score the edges in Equation \ref{eq:edge-scoring}. Note that for Polblogs, we use an identity matrix $I$ of size $N+k$ as our feature matrix throughout the framework. To demonstrate our efficiency, we choose $r = \{0.03, 0.07, 0.1, 0.15\}$.

{\bf Default target and base classes.} We select a target class, which has at least $100$ nodes in the test set. This ensures that we have enough attacker nodes even in a parsimonious setting -- very few attacker nodes w.r.t the target class nodes in test set. 
However, in Table \ref{tab:base-target-result}, we show that our model consistently outperforms other baselines with varying choices of base and target classes. 

{\bf Node classifier.} Throughout the paper, we consider GCN as the default node classifier. We also consider two other neural node classifiers -- GAT \citep{velivckovic2017graph} and GraphSAGE \citep{hamilton2017inductive}, and report their results in Table \ref{tab:gat}.

{\bf Hiding setup.}
We use our hiding framework as a pre-processing strategy. As described in Section \ref{hiding-framework}, the added pretend edges connect the attacker nodes to nodes in $V_b$. Once added, we mask them in the attack module to prevent them from further modification. On the other hand, we use CVAE \citep{sohn2015learning} to generate pretend features. However, CVAE does not guarantee discrete features; therefore, for discrete datasets, we define a threshold $\delta$ such that $PF_{i,j}=\mathbb{I}(PF'_{i,j}>\delta)$, where $PF'$ denotes the feature matrix obtained from CVAE. For both Cora and Citeseer, we set $\delta=0.1$ as it best mimics the average number of $1$s per vector in the original feature matrix $X$.

For Reddit, we do not need features to be discrete; hence, we skip the thresholding technique. In case of Polblogs, there is no feature matrix in the network; therefore, the pretend feature module is not required. For each network, we select the class containing the maximum number of nodes as our base class $l_b$ since injecting new nodes to the already large set ($V_b$) would make the attack unnoticeable.

\subsection{Performance Comparison}

We present results for two versions of our model -- \model\ and \model\ (hide). The former represents our core attacking strategy described till Section \ref{generator}, while the latter attaches the hiding framework on top of this. The experimental settings for both the versions are described Section \ref{sec:setup}. Table \ref{tab:class-results} shows the  accuracy of the node classifier on different datasets across varying budgets.

\begin{figure}[!t]
\centering
\includegraphics[width=0.7
\columnwidth]{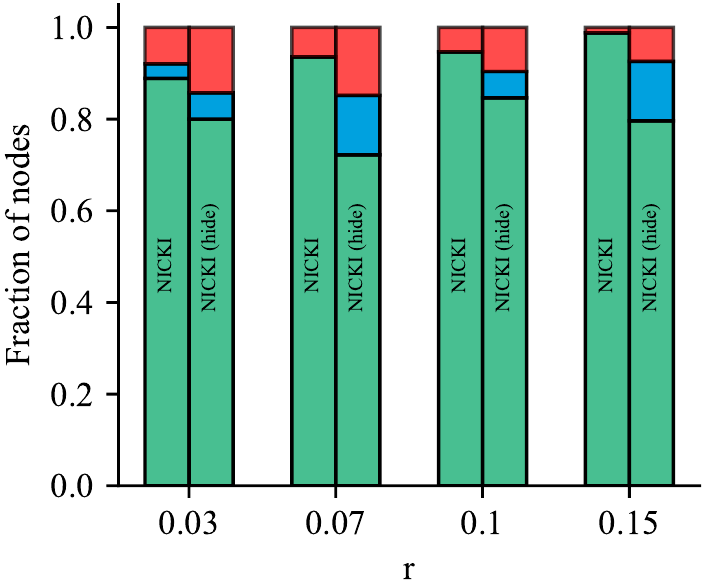}
  \caption{Fraction of target nodes in their respective predicted classes post network attack for Cora. {\color{greenf}Green} indicates nodes misclassified as the base class (our goal). {\color{bluef}Blue} indicates nodes correctly classified in target class, and {\color{red}Red} indicates nodes misclassified into a class other than base class.}
  \label{fig:hiding}
\end{figure}

\textbf{Cora, Citeseer and Polblogs.}
\model\ performs consistently better than every baseline by a considerable margin. The performance difference between \model\ and the best baseline is around $7-9\%$ for the lowest budget and goes as high as $23-35\%$ for a comparatively higher budget. AFGSM, as described in \citep{wang2020scalable}, uses a closed-form solution which is optimised to attack only a single target node. AFGSM scores each candidate perturbation w.r.t to misclassification of the target node. We adopt AFGSM to our multi-target setting by selecting perturbations that have a higher sum of scores for each node in the target class. The poor performance of AFGSM is indicative of its closed-form solution not being generalizable for attacking multiple target nodes. As expected, random and preferential models do not perform well; the misclassification increases as we increase the budget and in some cases gets better than AFGSM. NIPA, on the other hand, is a global attack, and hence, is better suited for our setting. The superior performance of NIPA compared to AFGSM on Cora and Polblogs can be attributed to the above fact.
\model\ (hide), as expected, has an inferior performance than \model. However, for Cora and Citeseer, it still outperforms other baselines by a significant margin. We can observe that the baselines do not necessarily show improvement in performance with increasing budget. It is attributed to the fact that these algorithms are not optimized for low-budget settings like ours. \model, on the other hand, although not fully consistent due to probabilistic nature of gradient based algorithms, shows an upward trend in misclassification with increasing budget. The parameter $\alpha$ in Equation \ref{total_loss} controls the extent of hiding, and in turn, balances the trade-off between better misclassification and better hiding. For Cora, we also observe the reclassification of target nodes after the attack. From Figure \ref{fig:hiding}, we can infer that majority of the nodes are reclassified as base class nodes. This stands as a testament to the working of \model\ towards fulfilling our objective defined in Equation \ref{eq:obj_ours}.

\textbf{Reddit.} For the lower budget at $r=0.03$, AFGSM performs marginally better than \model. However, as the budget increases, \model\ outperforms AFGSM by a significant margin (around $7\%$). AFGSM only generates discrete features, whereas Reddit is a continuous dataset. We still evaluate on discrete features to allow AFGSM to reach its full-attack potential, which we think might be the reason for its superior performance in some cases. NIPA gives random features to attacker nodes which doesn't work well for Reddit, given its continuous features. \model\ (hide) performs better than random and preferential but is not able to outperform AFGSM. We feel that the restriction on feature generation given by the \textit{hiding feature loss} does not allow our model to reach its attack potential in this case.

\textbf{Varying base and target classes.}\label{sec:varying} Table \ref{tab:class-results} shows the results for varying budgets but fixed target and base classes. To demonstrate the performance of our model for any given pair of target and base classes, we randomly select four such pairs for Cora and present the node classification accuracy in Table \ref{tab:base-target-result} for $r=0.07$.  We observe that our model outperforms the best baseline AFGSM across all choices.

\subsection{Hiding Efficiency}

\begin{figure}[!t]
\centering
\includegraphics[width=\columnwidth]{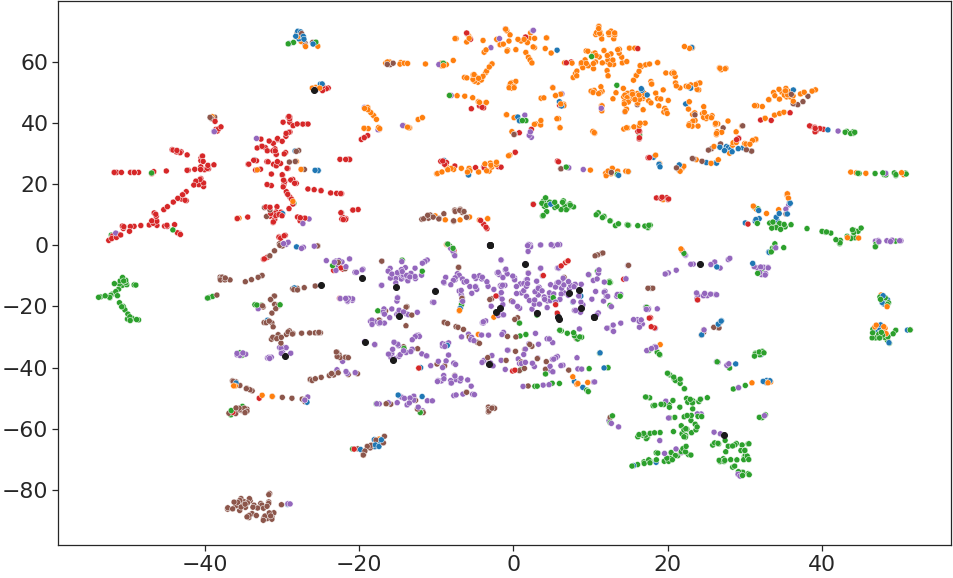}
\caption{2D t-SNE plot of Deepwalk embedding for Citeseer. Attacker nodes are black. Nodes in the base and target classes are {\color{violet}violet} and {\color{orange}orange}, respectively (see Figure \ref{appfig:embd} in Appendix for the feature embedding of Citeseer).}\label{citeseer-hide}
\end{figure}

We test the effectiveness of our hiding strategy by checking the closeness of both features and topological structure assumed by attacker nodes to the nodes in the base class. 

\begin{figure}
\centering
\includegraphics[width=0.7\columnwidth]{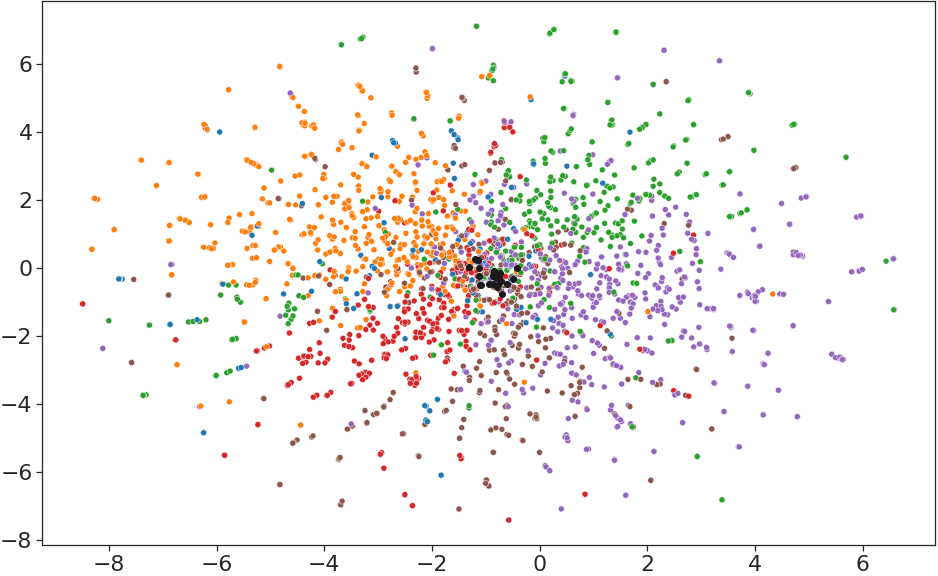}
\caption{2D t-SNE plot showing the feature embedding of Citeseer. Attacker nodes are black. Nodes in the base and target classes are {\color{violet}violet} and {\color{orange}orange}, respectively.} \label{appfig:embd}
\end{figure}

For features, we reduce dimensionality using t-SNE. Figure \ref{appfig:embd} displays the 2D t-SNE plot of representation of the nodes in Citeseer network. As visible, the attacker nodes (black) tend to be near the base class nodes (violet). Since the attack aims to bring the target and base class nodes together, the attacker nodes seem to bridge the nodes belonging to two classes. We can change the position of attacker nodes by tuning $\alpha$ in Equation \ref{total_loss}. 

For the topological structure, we use Deepwalk \citep{perozzi2014deepwalk} to obtain node embeddings, and then pass it to t-SNE (Figure \ref{citeseer-hide}). As visible, our injected attacker nodes seem to lay towards the pre-existing base class nodes' distribution and are good at picking up the topological structure of base class nodes, and thus can be easily camouflaged with them. 

Although removing the hiding framework increases the attack performance of \model\ in all cases, there is a trade-off with imperceptibility. Without the hiding framework, the injected nodes might be detected easily since there is no mechanism to ensure their camouflage. Also, our assumption that the injected nodes will be labelled as base-class nodes will become more feasible with hiding since it essentially tries to camouflage injected nodes with base-class nodes only. Better hiding strategy can be explored in the future to improve \model\ (hide)'s performance.

\begin{table}[!t]
    \centering
    \caption{Classification accuracy of target class nodes in test set across randomly selected base and target classes on Cora with $r=0.07$. Lower value indicates better attack.}
    \vspace{1mm}
    \scalebox{0.9}{
    \begin{tabular}{p{12mm}|p{12mm}|ccp{8mm}cc}
    \hline
        {\bf Base Class ID} & {\bf Target Class ID} & {\bf Clean} & {\bf \model\ } & {\bf \model\ (hide) } & {\bf AFGSM}& {\bf NIPA}\\\hline\hline
        5 & 0 & 0.818 & {\bf 0.507} & 0.727 & 0.785 & 0.746\\
        0 & 2 & 0.841 & {\bf 0.722} & 0.837 & 0.819 & 0.770\\
        1 & 6 & 0.821 & 0.805 & {\bf 0.793} & 0.818 & 0.805\\
        3 & 4 & 0.728 & {\bf 0.585} & 0.685 & 0.667 & 0.650\\
        \hline
    \end{tabular}}
    \label{tab:base-target-result}
\end{table}

\begin{table}[!th]
    \centering
    \caption{Comparison with Unnoticeable Evasion attacks. Interpretation of results is same as that of Table \ref{tab:class-results}}
    \scalebox{0.8}{
    \begin{tabular}{c|c|cccc}
    \hline
        \bf Dataset & \bf Method & \bf $r=0.03$ & \bf $r=0.07$ & \bf $r=0.10$ & \bf $r=0.15$\\\hline\hline
        \multirow{3}{*}{\makecell{ Cora \\ $[0.8189]$}} & GANI & 0.7966 & 0.7688 & 0.7966 & 0.7715 \\
        & AGIA + HAO & 0.7855 & 0.7855 & 0.8245 & 0.7910 \\
        & \model\ (hide) & {\color{red}0.7409} & {\color{red}0.7270} & {\color{red}0.7103} & {\color{red}0.7214} \\
        \hline
        \multirow{3}{*}{\makecell{ Citeseer \\ $[0.7395]$}} & GANI & 0.7368 & 0.7447 & {\color{red}0.7236} & 0.7368 \\
        & TDGIA + HAO & 0.7473 & 0.8184 & 0.8105 & 0.8289 \\
        & \model\ (hide) & {\color{red}0.7131} & {\color{red}0.7263} & 0.7316 & {\color{red}0.7316} \\
        \hline
    \end{tabular}}
    \label{tab:evasion-results}
\end{table}

\section{Comparison with Unnoticeable Evasion Attacks}\label{app:evasion}

We also compare our model with start-of-the-art unnoticeable evasion attacks -- HAO \cite{chen2022understanding} and GANI \cite{fang2022gani}. We evaluate them in a posison setting by simply training a vanilla GCN once on the attacked graph before evaluating target class accuracy. The idea is to emulate baselines in a real-world poison setting in which perturbations happen prior to training. For this purpose, we use two datasets -- Cora and Citeseer as both baselines scale well on them. Since HAO \cite{chen2022understanding} works as an unnoticeability inducing extension for other node-injection attacks, we use AGIA \cite{chen2022understanding} and TDGIA \cite{zou2021tdgia} as base attacks for Cora and Citeseer, respectively, since they worked best in \cite{chen2022understanding} for the respective datasets. Additionally, we convert GANI \cite{fang2022gani} into a targeted attack by optimising its genetic loss function only on target-class nodes. We present the results in Table \ref{tab:evasion-results}. To conduct a fair comparison in terms of unnoticeable attacks, we only include \model\ (hide) results in Table \ref{tab:evasion-results}.

\subsection{Performance Comparison}
We can observe from the Table \ref{tab:evasion-results} that \model\ (hide) outperforms in almost every case, showing that even in the presence of unnoticeability strategies, our model works better and compromises on accuracy much lesser than the other two baselines. GANI \cite{fang2022gani} still performs at-par with our model; however HAO performs significantly worse. This can be attributed to two reasons - (i) HAO works on a significantly higher feature perturbation budget which we limited by taking the top-k scores, leading to a reduction in performance. (ii) Evasion attacks in general perform worse in a poison setting as they are not optimised for training like \model\ is (Equation \ref{eq:loss-feat}).

\section{Transferability of Attack}\label{app:transfer}
In our approach, we choose GCN \citep{kipf2016semi} as our surrogate model and use it to evaluate the poisoned graph, the results of which are presented in Table \ref{tab:class-results}. However in a real-world scenario, an attacker has no prior knowledge of the classifier being used. Therefore, it only makes sense to verify the efficacy of our attack on other widely-used node classifiers, for which we select GAT \citep{velivckovic2017graph} and GraphSAGE \citep{hamilton2017inductive}. Note that since ours is a poison attack, we first train the obtained poisoned graph on the above mentioned classifiers and then evaluate the accuracy of target nodes. Results can be found in Table \ref{tab:transfer-attack}. We can observe that \model\ performs best in most cases followed by \model\ (hide); this shows that our attack can be effectively transferred to other GNN based classifiers.

\section{\model\ Against Defense}
\begin{table*}[!t]
    \centering
    \caption{Evaluation of attacks under robust defenses. Base, target class and the interpretation of results are same as that of Table \ref{tab:gat}.}\label{tab:defense}
    \scalebox{0.65}{
    \begin{tabular}{c|c|cc|cc}
    \hline
    & & \multicolumn{2}{c}GNNGuard & \multicolumn{2}{c}EGNNGuard \\
    \cline{3-6}
        \bf Dataset & \bf Method & \bf $r=0.03$ & \bf $r=0.10$ & \bf $r=0.03$ & \bf $r=0.10$\\\hline\hline
        \multirow{4}{*}{\makecell{ Cora \\ $[0.4484, 0.8161]$}} & GANI & 0.4568 & 0.4540 & 0.8161 & 0.8077  \\
        & AGIA + HAO & 0.3231 & 0.3454 & 0.7688 & 0.7883 \\
        & \model & 0.3788 & 0.1838 & 0.7493 & 0.4038 \\
        & \model\ (hide) & {\color{red}0.2228} & {\color{red}0.1559} & {\color{red}0.5515} & {\color{red}0.0863} \\
        \hline
        \multirow{4}{*}{\makecell{ Citeseer \\ $[0.8973, 0.7289]$}} & GANI & 0.8947 & 0.8789 & 0.7236 & 0.7078\\
        & TDGIA + HAO & 0.8973 & 0.9 & 0.7289 & 0.7421\\
        & \model & 0.6763 & {\color{red}0.3421} & 0.6868 & {\color{red}0.0473} \\
        & \model\ (hide) & {\color{red}0.3421} & {\color{red}0} & {\color{red}0.0473} & {\color{red}0} \\
        \hline
    \end{tabular}}
    \label{tab:robust-defense}
\end{table*}
In Table \ref{tab:gat}, we evaluate our attack on common GNN techniques - GAT and GraphSage. Realistically, institutions often employ more robust algorithms to prevent adversarial attacks from happening. Motivated by \cite{chen2022understanding} that homophily defenders work well against node-injection attacks, we also evaluate the efficacy of our attack against more robust GNN algorithms and homophily defenders - GNNGuard \cite{zhang2020gnnguard} and EGNNGuard \cite{chen2022understanding}.

$\blacksquare$ {\bf GNNGuard.}
This is a general algorithm used to improve robustness of any GNN model. GNNGuard assigns higher weightage to edges between similar nodes and prunes edges between dissimilar nodes. The new edges provide robust message passing to mitigate the effect of attacks.

$\blacksquare$ {\bf EGNNGuard.}
This is a more scalable version of GNNGuard based on similar ideology and tries to maintain homophily in the graph. It uses a different pruning method as it simply puts a threshold and removes edges between neighbour nodes with low similarity.

Table \ref{tab:robust-defense} shows that our attack is not affected by the above mentioned robust defenses. In fact, the accuracies are in most cases lower (better attack) than the ones in Table \ref{tab:evasion-results} in which we used a vanilla GCN. Since both these methods purge edges based on node similarities, we further plot pairwise node similarities and perform a homophily analysis to investigate further.

\subsection{Homophily Analysis}
\begin{figure}
    \centering
\begin{subfigure}{0.495\textwidth}
    \centering
    \includegraphics[width=\textwidth]{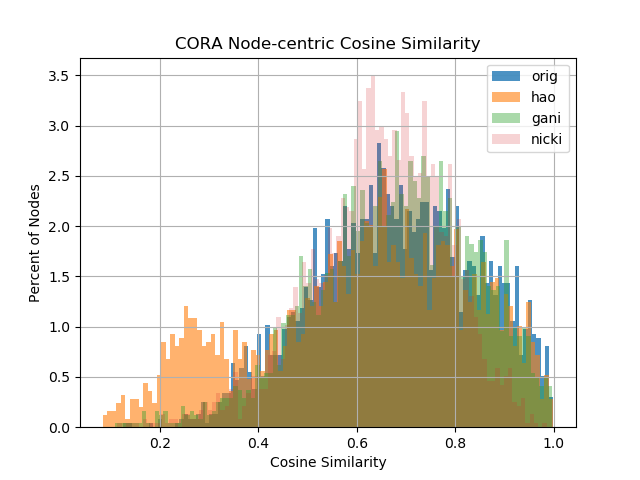}
    \caption{$r = 0.03$}
    \label{fig:node_sim1}
\end{subfigure}
\begin{subfigure}{0.495\textwidth}
    \centering
    \includegraphics[width=\textwidth]{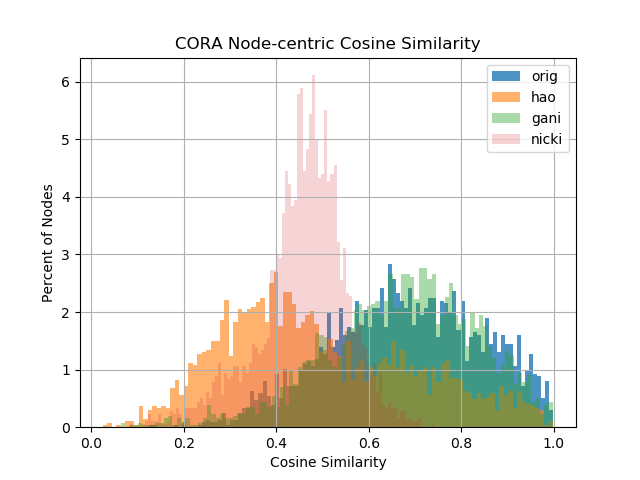}
    \caption{$r = 0.10$}
    \label{fig:node_sim2}
\end{subfigure}
\begin{subfigure}{0.495\textwidth}
    \centering
    \includegraphics[width=\textwidth]{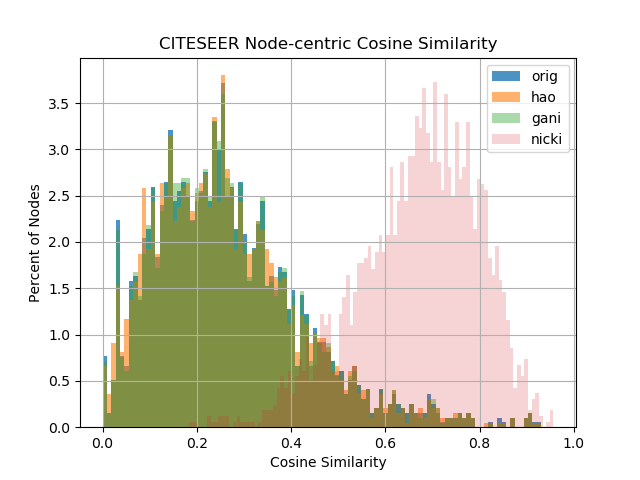}
    \caption{$r = 0.03$}
    \label{fig:node_sim3}
\end{subfigure}
\begin{subfigure}{0.495\textwidth}
    \centering
    \includegraphics[width=\textwidth]{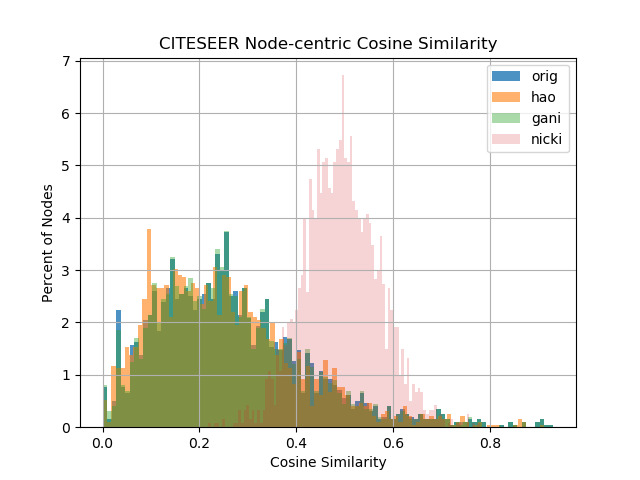}
    \caption{$r = 0.10$}
    \label{fig:node_sim4}
\end{subfigure}
    \caption{Plots to measure homophily change for different attacks on Cora and Citeseer. Note that for \model\ we presented results with hide. For Cora, HAO is used with AGIA and with TDGIA for Citeseer.}
    \label{fig:node_sim}
\end{figure}
We plot node similarity for the graphs generated by all the unnoticeable attacks on Cora and Citeseer on two different budgets in Figure \ref{fig:node_sim}. As expected at lower budgets, each plot seems to move towards the clean graph. GANI does an excellent job at keeping the distributions almost overlapped across both graphs and budgets. However, in case of HAO, the graph is left shifted for Cora, i.e., node injection is destroying homophily. Interestingly, \model\ (hide) either creates a peak (in Cora) or shifts right (in Citeseer). Shifting right is analogous to an increase in homophily which is unexpected by graph injection attacks and thus fools the defense model. Similarly peak creation leads to most node pairs centred around the same similarity value and makes it difficult to pick the notorious edges. In fact, we believe that because of this nature, the GNN defenses purge benign edges which leads to even better attack efficacy and sometimes complete disruption of the graph. We can verify these findings from Table \ref{tab:robust-defense} as well. 

\section{Post Attack Statistics}\label{app:post_attack}
\begin{table}[!t]
    \centering
    \caption{Statistics -- Gini coefficient (GC), distribution entropy (DE), power-law exponent (PE), and triangle count (TC) of different graphs after poisoning (\model;AFGSM). The values at $r=0$ indicate the statistics of the clean graph.}
    \scalebox{0.8}{
    \begin{tabular}{c|c|cccc}
    \hline
        & {\bf $r$} & {\bf GC} & {\bf DE}& {\bf PE}& {\bf TC}\\\hline\hline
        \multirow{6}{*}{\rotatebox{90}{Cora}} & 0 & 0.405 & 0.360 & 1.932 & 1630  \\\cline{2-6}
        & 0.03 & 0.403;0.405 & 0.361;0.361 & 1.926;1.927 & 1633;1636  \\
        & 0.07 & 0.402;0.405 & 0.364;0.364 & 1.918;1.921 & 1634;1642  \\
        & 0.1 & 0.401;0.406 & 0.365;0.366 & 1.911;1.917 & 1635;1640  \\
        & 0.15 & 0.399;0.408 & 0.367;0.369 & 1.902;1.911 & 1634;1640  \\
        \hline
        \multirow{6}{*}{\rotatebox{90}{Citeseer}} & 0 & 0.427 & 0.352 & 2.058 & 1083  \\\cline{2-6}
        & 0.03 & 0.424;0.428 & 0.354;0.352 & 2.047;2.062 & 1084;1083  \\
        & 0.07 & 0.424;0.430 & 0.357;0.354 & 2.037;2.058 & 1086;1083  \\
        & 0.10 & 0.424;0.431 & 0.360;0.355 & 2.031;2.054 & 1085;1086  \\
        & 0.15 & 0.429;0.432 & 0.355;0.357 & 2.044;2.045 & 1114;1086 \\
        \hline
        \multirow{6}{*}{\rotatebox{90}{Polblogs}} & 0 & 0.599 & 0.842 & 1.478 & 25649  \\\cline{2-6}
        & 0.03 & 0.593;0.587 & 0.848;0.852 & 1.472;1.466 & 25811;25653  \\
        & 0.07 & 0.588;0.576 & 0.856;0.858 & 1.465;1.458 & 25954;25661 \\
        & 0.10 & 0.568;0.569 & 0.867;0.862 & 1.446;1.453 & 25742;25665 \\
        & 0.15 & 0.558;0.560 & 0.873;0.869 & 1.437;1.446 & 25864;25681  \\
        \hline
        \multirow{6}{*}{\rotatebox{90}{Reddit}} & 0 & 0.479 & 0.439 & 1.693 & 6850  \\\cline{2-6}
        & 0.03 & 0.479;0.479 & 0.439;0.439 & 1.692;1.692 & 6850;6853  \\
        & 0.07 & 0.477;0.478 & 0.440;0.439 & 1.689;1.691 & 6855;6855 \\
        & 0.10 & 0.476;0.478 & 0.440;0.440 & 1.688;1.690 & 6856;6857 \\
        & 0.15 & 0.475;0.478 & 0.441;0.440 & 1.686;1.689 & 6855;6862  \\
        \hline
    \end{tabular}}
    \label{tab:result-post}
\end{table}
Following \cite{sun2020adversarial}, we obtain the poisoned graphs from \model\ and AFGSM, and compare them with their corresponding clean graphs by presenting some key network statistics in Table \ref{tab:result-post}.
The change in the degree distribution of a graph is considered as an important metric to measure {\em unnoticeability} of an attack \citep{zugner2018adversarial}. It is interesting to observe that for our attack, the power-law exponent of the poisoned graph is similar to that of the clean graph. This shows that our pre- and post-attack graphs have similar degree distributions; indicating unnoticeability. As expected, the triangle count increases as more nodes are injected. However, note that the increment is minor in most cases, which can be attributed to our very small budget, leading to less number of new triangles being induced by the attacker nodes. AFGSM shows a similar trend, which is interesting since \model\ is able to achieve much better results (Table \ref{tab:result-post}) while maintaining statistics similar to that of AFGSM. For the other statistics, we can observe that the numbers remain almost consistent even when the budget increases. Again, we impute this consistency to our minimal budget, which as we can see, helps in the overall unnoticeability of our attack.

\section{Conclusion}
In this paper, we introduced a novel problem -- class-specific network poisoning, which unlike existing methods, aims to poison a network in such a way that the nodes in the target class get misclassified. We addressed this problem by introducing \model, a novel attacking strategy that leverages an optimization-based approach to deteriorate the performance of a node classifier. Extensive experiments on four real-world datasets showed significant performance gain over four baselines in terms of misclassifying the nodes in the target class. We also showed that the attack graph  and clean graph look alike in terms of the several topological properties of the networks. We further empirically showed that the attacker nodes resemble benign nodes.

 \bibliographystyle{elsarticle-harv} 
 \bibliography{references}

\begin{thebibliography}{37}
\expandafter\ifx\csname natexlab\endcsname\relax\def\natexlab#1{#1}\fi
\providecommand{\url}[1]{\texttt{#1}}
\providecommand{\href}[2]{#2}
\providecommand{\path}[1]{#1}
\providecommand{\DOIprefix}{doi:}
\providecommand{\ArXivprefix}{arXiv:}
\providecommand{\URLprefix}{URL: }
\providecommand{\Pubmedprefix}{pmid:}
\providecommand{\doi}[1]{\href{http://dx.doi.org/#1}{\path{#1}}}
\providecommand{\Pubmed}[1]{\href{pmid:#1}{\path{#1}}}
\providecommand{\bibinfo}[2]{#2}
\ifx\xfnm\relax \def\xfnm[#1]{\unskip,\space#1}\fi
\bibitem[{Adamic and Glance(2005)}]{adamic2005political}
\bibinfo{author}{Adamic, L.A.}, \bibinfo{author}{Glance, N.},
  \bibinfo{year}{2005}.
\newblock \bibinfo{title}{The political blogosphere and the 2004 us election:
  divided they blog}, in: \bibinfo{booktitle}{Proceedings of the 3rd
  international workshop on Link discovery}, pp. \bibinfo{pages}{36--43}.
\bibitem[{Barab{\'a}si and Albert(1999)}]{barabasi1999emergence}
\bibinfo{author}{Barab{\'a}si, A.L.}, \bibinfo{author}{Albert, R.},
  \bibinfo{year}{1999}.
\newblock \bibinfo{title}{Emergence of scaling in random networks}.
\newblock \bibinfo{journal}{science} \bibinfo{volume}{286},
  \bibinfo{pages}{509--512}.
\bibitem[{Bojchevski and G{\"u}nnemann(2019)}]{bojchevski2019adversarial}
\bibinfo{author}{Bojchevski, A.}, \bibinfo{author}{G{\"u}nnemann, S.},
  \bibinfo{year}{2019}.
\newblock \bibinfo{title}{Adversarial attacks on node embeddings via graph
  poisoning}, in: \bibinfo{booktitle}{ICML}, pp. \bibinfo{pages}{695--704}.
\bibitem[{Chen et~al.(2022)Chen, Yang, Zhang, Ma, Liu, Han and
  Cheng}]{chen2022understanding}
\bibinfo{author}{Chen, Y.}, \bibinfo{author}{Yang, H.}, \bibinfo{author}{Zhang,
  Y.}, \bibinfo{author}{Ma, K.}, \bibinfo{author}{Liu, T.},
  \bibinfo{author}{Han, B.}, \bibinfo{author}{Cheng, J.}, \bibinfo{year}{2022}.
\newblock \bibinfo{title}{Understanding and improving graph injection attack by
  promoting unnoticeability}.
\newblock \bibinfo{journal}{arXiv preprint arXiv:2202.08057} .
\bibitem[{Dai et~al.(2018)Dai, Li, Tian, Huang, Wang, Zhu and
  Song}]{dai2018adversarial}
\bibinfo{author}{Dai, H.}, \bibinfo{author}{Li, H.}, \bibinfo{author}{Tian,
  T.}, \bibinfo{author}{Huang, X.}, \bibinfo{author}{Wang, L.},
  \bibinfo{author}{Zhu, J.}, \bibinfo{author}{Song, L.}, \bibinfo{year}{2018}.
\newblock \bibinfo{title}{Adversarial attack on graph structured data}, in:
  \bibinfo{booktitle}{ICML}, pp. \bibinfo{pages}{1115--1124}.
\bibitem[{Erdos et~al.(1960)Erdos, R{\'e}nyi et~al.}]{erdos1960evolution}
\bibinfo{author}{Erdos, P.}, \bibinfo{author}{R{\'e}nyi, A.}, et~al.,
  \bibinfo{year}{1960}.
\newblock \bibinfo{title}{On the evolution of random graphs}.
\newblock \bibinfo{journal}{Publ. Math. Inst. Hung. Acad. Sci}
  \bibinfo{volume}{5}, \bibinfo{pages}{17--60}.
\bibitem[{Fan et~al.(2019)Fan, Ma, Li, He, Zhao, Tang and Yin}]{fan2019graph}
\bibinfo{author}{Fan, W.}, \bibinfo{author}{Ma, Y.}, \bibinfo{author}{Li, Q.},
  \bibinfo{author}{He, Y.}, \bibinfo{author}{Zhao, E.}, \bibinfo{author}{Tang,
  J.}, \bibinfo{author}{Yin, D.}, \bibinfo{year}{2019}.
\newblock \bibinfo{title}{Graph neural networks for social recommendation}, in:
  \bibinfo{booktitle}{WWW}, pp. \bibinfo{pages}{417--426}.
\bibitem[{Fang et~al.(2022)Fang, Wen, Wu, Xuan, Zheng and Tse}]{fang2022gani}
\bibinfo{author}{Fang, J.}, \bibinfo{author}{Wen, H.}, \bibinfo{author}{Wu,
  J.}, \bibinfo{author}{Xuan, Q.}, \bibinfo{author}{Zheng, Z.},
  \bibinfo{author}{Tse, C.K.}, \bibinfo{year}{2022}.
\newblock \bibinfo{title}{Gani: Global attacks on graph neural networks via
  imperceptible node injections}.
\newblock \bibinfo{journal}{arXiv preprint arXiv:2210.12598} .
\bibitem[{Giles et~al.(1998)Giles, Bollacker and
  Lawrence}]{10.1145/276675.276685}
\bibinfo{author}{Giles, C.L.}, \bibinfo{author}{Bollacker, K.D.},
  \bibinfo{author}{Lawrence, S.}, \bibinfo{year}{1998}.
\newblock \bibinfo{title}{Citeseer: An automatic citation indexing system}, in:
  \bibinfo{booktitle}{JCDL}, p. \bibinfo{pages}{89–98}.
\bibitem[{Goodfellow et~al.(2015)Goodfellow, Shlens and
  Szegedy}]{goodfellow2014explaining}
\bibinfo{author}{Goodfellow, I.J.}, \bibinfo{author}{Shlens, J.},
  \bibinfo{author}{Szegedy, C.}, \bibinfo{year}{2015}.
\newblock \bibinfo{title}{Explaining and harnessing adversarial examples}, in:
  \bibinfo{booktitle}{ICLR}.
\bibitem[{Grover and Leskovec(2016)}]{grover2016node2vec}
\bibinfo{author}{Grover, A.}, \bibinfo{author}{Leskovec, J.},
  \bibinfo{year}{2016}.
\newblock \bibinfo{title}{node2vec: Scalable feature learning for networks},
  in: \bibinfo{booktitle}{SIGKDD}, pp. \bibinfo{pages}{855--864}.
\bibitem[{Hamilton et~al.(2017)Hamilton, Ying and
  Leskovec}]{hamilton2017inductive}
\bibinfo{author}{Hamilton, W.L.}, \bibinfo{author}{Ying, R.},
  \bibinfo{author}{Leskovec, J.}, \bibinfo{year}{2017}.
\newblock \bibinfo{title}{Inductive representation learning on large graphs},
  in: \bibinfo{booktitle}{NIPS}, pp. \bibinfo{pages}{1025--1035}.
\bibitem[{Jia and Liang(2017)}]{jia2017adversarial}
\bibinfo{author}{Jia, R.}, \bibinfo{author}{Liang, P.}, \bibinfo{year}{2017}.
\newblock \bibinfo{title}{Adversarial examples for evaluating reading
  comprehension systems}, in: \bibinfo{booktitle}{EMNLP}, pp.
  \bibinfo{pages}{2021--2031}.
\bibitem[{Jiang et~al.(2020)Jiang, Li, Zhang, Wang, Wang, Yuan and
  Wei}]{jiang2020drug}
\bibinfo{author}{Jiang, M.}, \bibinfo{author}{Li, Z.}, \bibinfo{author}{Zhang,
  S.}, \bibinfo{author}{Wang, S.}, \bibinfo{author}{Wang, X.},
  \bibinfo{author}{Yuan, Q.}, \bibinfo{author}{Wei, Z.}, \bibinfo{year}{2020}.
\newblock \bibinfo{title}{Drug--target affinity prediction using graph neural
  network and contact maps}.
\newblock \bibinfo{journal}{RSC Advances} \bibinfo{volume}{10},
  \bibinfo{pages}{20701--20712}.
\bibitem[{Jin et~al.(2019)Jin, Liu, Li, He and Zhang}]{jin2019graph}
\bibinfo{author}{Jin, D.}, \bibinfo{author}{Liu, Z.}, \bibinfo{author}{Li, W.},
  \bibinfo{author}{He, D.}, \bibinfo{author}{Zhang, W.}, \bibinfo{year}{2019}.
\newblock \bibinfo{title}{Graph convolutional networks meet markov random
  fields: Semi-supervised community detection in attribute networks}, in:
  \bibinfo{booktitle}{AAAI}, pp. \bibinfo{pages}{152--159}.
\bibitem[{Kipf and Welling(2016)}]{kipf2016variational}
\bibinfo{author}{Kipf, T.N.}, \bibinfo{author}{Welling, M.},
  \bibinfo{year}{2016}.
\newblock \bibinfo{title}{Variational graph auto-encoders}.
\newblock \bibinfo{journal}{CoRR} \bibinfo{volume}{abs/1611.07308}.
\bibitem[{Kipf and Welling(2017)}]{kipf2016semi}
\bibinfo{author}{Kipf, T.N.}, \bibinfo{author}{Welling, M.},
  \bibinfo{year}{2017}.
\newblock \bibinfo{title}{Semi-supervised classification with graph
  convolutional networks}, in: \bibinfo{booktitle}{ICLR}, pp.
  \bibinfo{pages}{1--14}.
\bibitem[{Li et~al.(2020)Li, Jin, Xu and Tang}]{li2020deeprobust}
\bibinfo{author}{Li, Y.}, \bibinfo{author}{Jin, W.}, \bibinfo{author}{Xu, H.},
  \bibinfo{author}{Tang, J.}, \bibinfo{year}{2020}.
\newblock \bibinfo{title}{Deeprobust: A pytorch library for adversarial attacks
  and defenses}.
\newblock \bibinfo{journal}{arXiv preprint arXiv:2005.06149} .
\bibitem[{McCallum et~al.(2000)McCallum, Nigam, Rennie and
  Seymore}]{mccallum2000automating}
\bibinfo{author}{McCallum, A.K.}, \bibinfo{author}{Nigam, K.},
  \bibinfo{author}{Rennie, J.}, \bibinfo{author}{Seymore, K.},
  \bibinfo{year}{2000}.
\newblock \bibinfo{title}{Automating the construction of internet portals with
  machine learning}.
\newblock \bibinfo{journal}{Information Retrieval} \bibinfo{volume}{3},
  \bibinfo{pages}{127--163}.
\bibitem[{Pennington et~al.(2014)Pennington, Socher and
  Manning}]{pennington2014glove}
\bibinfo{author}{Pennington, J.}, \bibinfo{author}{Socher, R.},
  \bibinfo{author}{Manning, C.D.}, \bibinfo{year}{2014}.
\newblock \bibinfo{title}{Glove: Global vectors for word representation}, in:
  \bibinfo{booktitle}{EMNLP}, pp. \bibinfo{pages}{1532--1543}.
\bibitem[{Perozzi et~al.(2014)Perozzi, Al-Rfou and
  Skiena}]{perozzi2014deepwalk}
\bibinfo{author}{Perozzi, B.}, \bibinfo{author}{Al-Rfou, R.},
  \bibinfo{author}{Skiena, S.}, \bibinfo{year}{2014}.
\newblock \bibinfo{title}{Deepwalk: Online learning of social representations},
  in: \bibinfo{booktitle}{SIGKDD}, pp. \bibinfo{pages}{701--710}.
\bibitem[{Shafahi et~al.(2018)Shafahi, Huang, Najibi, Suciu, Studer, Dumitras
  and Goldstein}]{shafahi2018poison}
\bibinfo{author}{Shafahi, A.}, \bibinfo{author}{Huang, W.R.},
  \bibinfo{author}{Najibi, M.}, \bibinfo{author}{Suciu, O.},
  \bibinfo{author}{Studer, C.}, \bibinfo{author}{Dumitras, T.},
  \bibinfo{author}{Goldstein, T.}, \bibinfo{year}{2018}.
\newblock \bibinfo{title}{Poison frogs! targeted clean-label poisoning attacks
  on neural networks}, in: \bibinfo{booktitle}{NIPS}, p.
  \bibinfo{pages}{6106–6116}.
\bibitem[{Sohn et~al.(2015)Sohn, Lee and Yan}]{sohn2015learning}
\bibinfo{author}{Sohn, K.}, \bibinfo{author}{Lee, H.}, \bibinfo{author}{Yan,
  X.}, \bibinfo{year}{2015}.
\newblock \bibinfo{title}{Learning structured output representation using deep
  conditional generative models}.
\newblock \bibinfo{journal}{NIPS} \bibinfo{volume}{28},
  \bibinfo{pages}{3483--3491}.
\bibitem[{Sun et~al.(2020)Sun, Wang, Tang, Hsieh and
  Honavar}]{sun2020adversarial}
\bibinfo{author}{Sun, Y.}, \bibinfo{author}{Wang, S.}, \bibinfo{author}{Tang,
  X.}, \bibinfo{author}{Hsieh, T.Y.}, \bibinfo{author}{Honavar, V.},
  \bibinfo{year}{2020}.
\newblock \bibinfo{title}{Adversarial attacks on graph neural networks via node
  injections: A hierarchical reinforcement learning approach}, in:
  \bibinfo{booktitle}{The WebConf}, pp. \bibinfo{pages}{673--683}.
\bibitem[{Tao et~al.(2021)Tao, Cao, Shen, Huang, Wu and Cheng}]{tao2021single}
\bibinfo{author}{Tao, S.}, \bibinfo{author}{Cao, Q.}, \bibinfo{author}{Shen,
  H.}, \bibinfo{author}{Huang, J.}, \bibinfo{author}{Wu, Y.},
  \bibinfo{author}{Cheng, X.}, \bibinfo{year}{2021}.
\newblock \bibinfo{title}{Single node injection attack against graph neural
  networks}, in: \bibinfo{booktitle}{CIKM}, pp. \bibinfo{pages}{1794--1803}.
\bibitem[{Tao et~al.(2022)Tao, Cao, Shen, Wu, Hou and
  Cheng}]{tao2022adversarial}
\bibinfo{author}{Tao, S.}, \bibinfo{author}{Cao, Q.}, \bibinfo{author}{Shen,
  H.}, \bibinfo{author}{Wu, Y.}, \bibinfo{author}{Hou, L.},
  \bibinfo{author}{Cheng, X.}, \bibinfo{year}{2022}.
\newblock \bibinfo{title}{Adversarial camouflage for node injection attack on
  graphs}.
\newblock \bibinfo{journal}{arXiv preprint arXiv:2208.01819} .
\bibitem[{Veli{\v{c}}kovi{\'c} et~al.(2017)Veli{\v{c}}kovi{\'c}, Cucurull,
  Casanova, Romero, Lio and Bengio}]{velivckovic2017graph}
\bibinfo{author}{Veli{\v{c}}kovi{\'c}, P.}, \bibinfo{author}{Cucurull, G.},
  \bibinfo{author}{Casanova, A.}, \bibinfo{author}{Romero, A.},
  \bibinfo{author}{Lio, P.}, \bibinfo{author}{Bengio, Y.},
  \bibinfo{year}{2017}.
\newblock \bibinfo{title}{Graph attention networks}.
\newblock \bibinfo{journal}{arXiv preprint arXiv:1710.10903} .
\bibitem[{Wang et~al.(2020)Wang, Luo, Suya, Li, Yang and
  Zheng}]{wang2020scalable}
\bibinfo{author}{Wang, J.}, \bibinfo{author}{Luo, M.}, \bibinfo{author}{Suya,
  F.}, \bibinfo{author}{Li, J.}, \bibinfo{author}{Yang, Z.},
  \bibinfo{author}{Zheng, Q.}, \bibinfo{year}{2020}.
\newblock \bibinfo{title}{Scalable attack on graph data by injecting vicious
  nodes}.
\newblock \bibinfo{journal}{Data Mining and Knowledge Discovery}
  \bibinfo{volume}{34}, \bibinfo{pages}{1363--1389}.
\bibitem[{Wang et~al.(2018)Wang, Cheng, Eaton, Hsieh and Wu}]{wang2018attack}
\bibinfo{author}{Wang, X.}, \bibinfo{author}{Cheng, M.},
  \bibinfo{author}{Eaton, J.}, \bibinfo{author}{Hsieh, C.J.},
  \bibinfo{author}{Wu, F.}, \bibinfo{year}{2018}.
\newblock \bibinfo{title}{Attack graph convolutional networks by adding fake
  nodes}.
\newblock \bibinfo{journal}{arXiv preprint arXiv:1810.10751} .
\bibitem[{Xie and Ermon(2019)}]{xie2019reparameterizable}
\bibinfo{author}{Xie, S.M.}, \bibinfo{author}{Ermon, S.}, \bibinfo{year}{2019}.
\newblock \bibinfo{title}{Reparameterizable subset sampling via continuous
  relaxations}.
\newblock \bibinfo{journal}{arXiv preprint arXiv:1901.10517} .
\bibitem[{Xu et~al.(2019a)Xu, Shen, Cao, Cen and Cheng}]{xu2020graph}
\bibinfo{author}{Xu, B.}, \bibinfo{author}{Shen, H.}, \bibinfo{author}{Cao,
  Q.}, \bibinfo{author}{Cen, K.}, \bibinfo{author}{Cheng, X.},
  \bibinfo{year}{2019}a.
\newblock \bibinfo{title}{Graph convolutional networks using heat kernel for
  semi-supervised learning}, in: \bibinfo{booktitle}{IJCAI}, p.
  \bibinfo{pages}{1928–1934}.
\bibitem[{Xu et~al.(2019b)Xu, Shen, Cao, Qiu and Cheng}]{xu2019graph}
\bibinfo{author}{Xu, B.}, \bibinfo{author}{Shen, H.}, \bibinfo{author}{Cao,
  Q.}, \bibinfo{author}{Qiu, Y.}, \bibinfo{author}{Cheng, X.},
  \bibinfo{year}{2019}b.
\newblock \bibinfo{title}{Graph wavelet neural network}.
\newblock \bibinfo{journal}{arXiv preprint arXiv:1904.07785} .
\bibitem[{Ying et~al.(2018)Ying, He, Chen, Eksombatchai, Hamilton and
  Leskovec}]{ying2018graph}
\bibinfo{author}{Ying, R.}, \bibinfo{author}{He, R.}, \bibinfo{author}{Chen,
  K.}, \bibinfo{author}{Eksombatchai, P.}, \bibinfo{author}{Hamilton, W.L.},
  \bibinfo{author}{Leskovec, J.}, \bibinfo{year}{2018}.
\newblock \bibinfo{title}{Graph convolutional neural networks for web-scale
  recommender systems}, in: \bibinfo{booktitle}{SIGKDD}, pp.
  \bibinfo{pages}{974--983}.
\bibitem[{Zhang and Zitnik(2020)}]{zhang2020gnnguard}
\bibinfo{author}{Zhang, X.}, \bibinfo{author}{Zitnik, M.},
  \bibinfo{year}{2020}.
\newblock \bibinfo{title}{Gnnguard: Defending graph neural networks against
  adversarial attacks}.
\newblock \bibinfo{journal}{Advances in neural information processing systems}
  \bibinfo{volume}{33}, \bibinfo{pages}{9263--9275}.
\bibitem[{Zou et~al.(2021)Zou, Zheng, Dong, Guan, Kharlamov, Lu and
  Tang}]{zou2021tdgia}
\bibinfo{author}{Zou, X.}, \bibinfo{author}{Zheng, Q.}, \bibinfo{author}{Dong,
  Y.}, \bibinfo{author}{Guan, X.}, \bibinfo{author}{Kharlamov, E.},
  \bibinfo{author}{Lu, J.}, \bibinfo{author}{Tang, J.}, \bibinfo{year}{2021}.
\newblock \bibinfo{title}{Tdgia: Effective injection attacks on graph neural
  networks}.
\newblock \bibinfo{journal}{arXiv preprint arXiv:2106.06663} .
\bibitem[{Z{\"u}gner et~al.(2018)Z{\"u}gner, Akbarnejad and
  G{\"u}nnemann}]{zugner2018adversarial}
\bibinfo{author}{Z{\"u}gner, D.}, \bibinfo{author}{Akbarnejad, A.},
  \bibinfo{author}{G{\"u}nnemann, S.}, \bibinfo{year}{2018}.
\newblock \bibinfo{title}{Adversarial attacks on neural networks for graph
  data}, in: \bibinfo{booktitle}{SIGKDD}, pp. \bibinfo{pages}{2847--2856}.
\bibitem[{Z{\"u}gner and G{\"u}nnemann(2019)}]{zugner2019adversarial}
\bibinfo{author}{Z{\"u}gner, D.}, \bibinfo{author}{G{\"u}nnemann, S.},
  \bibinfo{year}{2019}.
\newblock \bibinfo{title}{Adversarial attacks on graph neural networks via meta
  learning}.
\newblock \bibinfo{journal}{arXiv preprint arXiv:1902.08412} .

\end{thebibliography}






\end{document}